\documentclass[journal]{IEEEtran}
\ifCLASSINFOpdf
\else
\fi
\usepackage{graphicx}
\usepackage{subfigure}
\usepackage{latexsym}
\usepackage{amsmath}
\usepackage{color}
\usepackage{algorithm}
\usepackage{algorithmic}

\graphicspath{{./figures/}}
\gdef\etal{et al.}

\begin{document}
\title{ENFT: Efficient Non-Consecutive Feature Tracking for Robust Structure-from-Motion}

\author{Guofeng~Zhang, Haomin Liu, Zilong Dong, Jiaya Jia, Tien-Tsin Wong, and Hujun Bao\IEEEcompsocitemizethanks{\IEEEcompsocthanksitem Guofeng Zhang, Haomin Liu, Zilong Dong and Hujun Bao are with the State Key Lab of CAD\&CG, Zhejiang University. Guofeng Zhang and Hujun Bao are also affiliated with Innovation Joint Research Center for Cyber-Physical-Society System, Zhejiang University. Email: \{zhangguofeng, zldong, bao\}@cad.zju.edu.cn, 172753015@qq.com. Corresponding authors: Guofeng Zhang and Hujun Bao.
		\IEEEcompsocthanksitem Jiaya Jia and Tien-Tsin Wong are with The Chinese University of Hong Kong. Tien-Tsin Wong is also affiliated with Shenzhen Research Institute, The Chinese University of Hong Kong. Email: \{leojia, ttwong\}@cse.cuhk.edu.hk}}

\markboth{}%
{Shell \MakeLowercase{\textit{et al.}}: Bare Demo of IEEEtran.cls for Journals}



\maketitle

\begin{abstract}
Structure-from-motion (SfM) largely relies on feature tracking. In image sequences, if
disjointed tracks caused by objects moving in and out of the field of view, occasional
occlusion, or image noise, are not handled well, corresponding SfM could be affected.
This problem becomes severer for large-scale scenes, which typically requires to capture
multiple sequences to cover the whole scene. In this paper, we propose an efficient
non-consecutive feature tracking (ENFT) framework to match interrupted tracks distributed
in different subsequences or even in different videos. Our framework consists of steps of
solving the feature `dropout' problem when indistinctive structures, noise or large image
distortion exists, and of rapidly recognizing and joining common features located in
different subsequences. In addition, we contribute an effective segment-based
coarse-to-fine SfM algorithm for robustly handling large datasets. Experimental results
on challenging video data demonstrate the effectiveness of the proposed system.
\end{abstract}


% \fi
\IEEEpeerreviewmaketitle

\section{Introduction}
Large-scale 3D
reconstruction~\cite{PollefeysNFAMCEGKMSSTWYSYWT08,FurukawaCSS10,FrahmGGJRWJDCL10}
finds many practical applications. It primarily relies on SfM
algorithms~\cite{Hartley2004,SnavelySS06,ZhangQHWHB07,AgarwalSSSS09,CrandallOSH11}
to firstly estimate sparse 3D features and camera poses given the input of video or image collections.

Compared to images, videos contain denser geometrical and structural information, and are
the main source of SfM in the movie, video and commercial industry. A common strategy for
video SfM estimation is by employing feature point tracking~\cite{TomasiK91,Lowe04},
which takes care of the temporal relationship among frames. It is also a basic tool for
solving a variety of computer vision problems, such as camera tracking, video matching,
and object recognition.

In this paper, we address two critical problems for feature point tracking, which could
handicap SfM especially for large-scale scene modeling. The first problem is the
vulnerability of feature tracking to object occlusions, illumination change, noise, and
large motion, which easily causes occasional feature drop-out and distraction. This
problem makes robust feature tracking from long sequences challenging.

The other problem is the inability of sequential feature tracking to cope with feature
matching over non-consecutive subsequences. A typical scenario is that the tracked object
moves out and then re-enters the field-of-view, which yields two discontinuous
subsequences containing the same object. Although there are common features in the two
subsequences, they are difficult to be matched/included in a single track using conventional
tracking methods. Addressing this issue can alleviate the drift problem of SfM, which
benefits high-quality 3D reconstruction. A na\"{\i}ve solution is to exhaustively search
all features, which could consume much computation since many temporally far away frames
simply share no content.

\begin{figure}[t]
\centering
\includegraphics[width=1.0\linewidth]{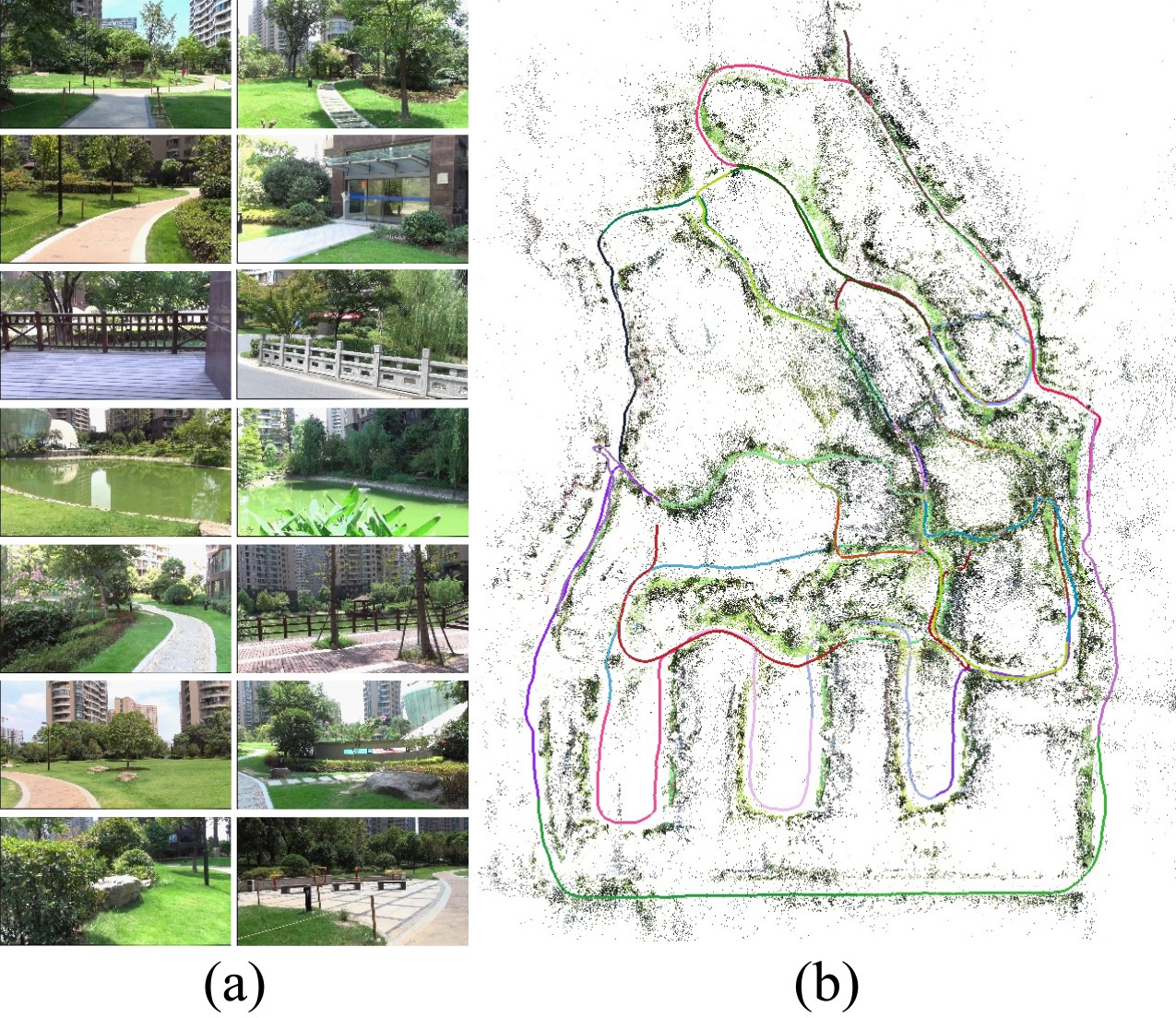}
\caption{A large-scale ``Garden" example. (a) Snapshots of the input videos. (b) With the
matched feature tracks, we register the recovered 3D points and camera trajectories in a large-scale 3D
system. Camera trajectories are differently color-coded. } \label{fig:garden}
\end{figure}

We propose an efficient non-consecutive feature tracking (ENFT) framework which can
effectively address the above problems in two phases -- that is, \emph{consecutive point
tracking} and \emph{non-consecutive track matching}. We demonstrate their significance
for SfM using challenging sequence data. \emph{Consecutive point tracking} detects and
matches invariant features in consecutive frames. A matching strategy is proposed to
greatly increase the matching rate and extend lifetime of the tracks. Then in {\em
non-consecutive track matching}, by rapidly computing a matching matrix, a set of
disjoint subsequences with overlapping content can be detected. Common feature tracks
scattered over these subsequences can also be reliably matched.

Our ENFT method reduces estimation errors for long loopback sequences. Given limited
memory, it is generally intractable to use global bundle adjustment to refine camera
poses and 3D points for very long sequences. Iteratively applying local bundle adjustment
is difficult to effectively distribute estimation errors to all frames. We address this
issue by adopting a segment-based coarse-to-fine SfM algorithm, which globally optimizes
structure and motion with limited memory.

Based on our ENFT algorithm and segment-based coarse-to-fine estimation scheme, we
present the SfM system {\it ENFT-SFM}, which can effectively handle long loopback
sequences and even multiple sequences. Fig.~\ref{fig:garden} shows an example containing
6 sequences with about $95,476$ frames in total in a large-scale scene. Our system splits
them to 37 shorter sequences, quickly computes many long and accurate feature tracks,
efficiently estimates camera trajectories in different sequences, and finally registers
them in a unified 3D system, as shown in Fig.~\ref{fig:garden}(b). The whole process only
takes about 90 minutes~(excluding I/O) on a desktop PC, i.e., 17.7 FPS on average. Our
supplementary
video\footnote{http://www.cad.zju.edu.cn/home/gfzhang/projects/tracking\\/featuretracking/ENFT-video.wmv}
contains the complete result.

Compared to our prior work~\cite{ZhangDJWB10}, in this paper, we make a number of
modifications to improve robustness and efficiency. Particularly, we improve the
second-pass matching by formulating it as minimizing an energy function incorporating two
geometric constraints, which not only boosts the matching accuracy but also reduces
computation. The non-consecutive track matching algorithm is re-designed to perform
feature matching and match-matrix update together. It is less sensitive to initialization
and reduces the matching time. Finally, we propose a novel segment-based
coarse-to-fine SfM method, which performs efficient global optimization for large data
with only limited memory.

\section{Related Work}
We review feature tracking and large-scale SfM methods in this
section.

\subsection{Feature Matching and Tracking}
For video tracking, sequential matchers are used for establishing correspondences between
consecutive frames. Kanade-Lucas-Tomasi (KLT) tracker~\cite{Lucas81,TomasiK91} is
widely used for small baseline matching. Other methods detect image features and match
them considering local image patches~\cite{NisterNB04,RoyerLDL07} or advanced
descriptors~\cite{Lowe04,MikolajczykS05,MatasCUP04,BayETG08}.

Both the KLT tracker and invariant feature algorithms depend on modeling feature
appearance, and can be distracted by occlusion, similar structures, and noise. Generally,
sequential matchers are difficult to match non-consecutive frames under image transformation.
Scale-invariant feature detection and matching algorithms~\cite{Lowe04,BayETG08} are
effective in matching images with large transformation. But they generally produce many
short tracks in consecutive point tracking due primarily to the global indistinctiveness
and feature dropout problems. In addition, invariant features are relatively sensitive to
perspective distortion. Although variations, such as ASIFT~\cite{MichelYu09}, can improve
matching performance under substantial viewpoint change, computation overhead increases
owing to exhaustive viewpoint simulation. Cordes~\etal~\cite{Cordes2011} proposed a
memory-based tracking method to extend feature trajectories by matching each frame to its
neighbors. However, if an object re-enters the field-of-view after a long period of time,
the size of neighborhood windows has to be very large. Besides, multiple-video setting
was not discussed. In contrast, our method can not only extend track lifetime but also efficiently match
common feature tracks in different subsequences by iteratively matching overlapping frame
pairs and refining match matrix. The computation complexity is linear to the number of
overlapping frame pairs.

There are methods using invariant features for object and location recognition in
images/videos~\cite{SivicZ03,SchaffalitzkyZ03,HoN07,Schindler2007,IrscharaZFB09}. These
methods typically use bag-of-words techniques to perform global localization and
loop-closure detection in an image classification framework. 
Nist{\'e}r and Stew{\'e}nius~\cite{NisterS06} proposed using a hierachical $k$-means algorithm to
construct a vocabulary tree with feature descriptors, which can be used for large-scale image retrieval and location recognition.
Cummins and Newman~\cite{CumminsN08} proposed a probabilistic approach called FAB-MAP for location recognition and online loop closure detection, which models the world as a set of locations and computes the probability of belonging to previously visited locations for each input image. Later, they proposed using a sparse approximation for large scale location recognition~\cite{CumminsN11}. However, FAB-MAP assumes the neighboring locations are not too close, so might perform less satisfyingly if we simply input a normal video sequence.
In addition, existing methods generally divide the location recognition and
non-consecutive feature matching into two separated phases
~\cite{LimFP11,ClippLFP10,EadeD08,HavlenaTP10}. Because the match
matrix by bag-of-words only roughly reflects the match confidence,
completely trusting it may lose many common features. In this
paper, we introduce a novel strategy where the match matrix can be
refined and updated along with non-consecutive feature matching.
Our method can reliably and efficiently match the common features
even with a coarse match matrix.

Engels et al. \cite{EngelsFN08} proposed integrating wide-baseline local features with
the tracked ones to improve SfM. The method creates small and independent submaps and
links them via feature recognition. This approach also cannot produce many long and
accurate point tracks. Short tracks are not enough for drift-free SfM estimation. In
comparison, our method is effective in high-quality point track estimation. We also
address the ubiquitous nondistinctive feature matching problem in dense frames. Similar
to the scheme of \cite{GrabnerB05}, we utilize track descriptors, instead of the feature
descriptors, to reduce computation redundancy.

Wu et al.~\cite{WuCLFP08} proposed using dense 3D geometry information to extend SIFT
features. In contrast, our method only uses sparse matches to estimate a set of
homographies to represent scene motion, which also handles viewpoint change. It is
general since geometry is not required.

\subsection{Large-Scale Structure-from-Motion}
State-of-the-art large-scale SfM methods can handle millions of images on a single PC in
one day~\cite{FrahmGGJRWJDCL10}. To this end, large image data are separated into a
number of independent submaps, each is optimized independently.
Steedly~\etal~\cite{SteedlyED03} proposed a partitioning approach to decompose a
large-scale optimization into multiple better-conditioned subproblems.
Clemente~\etal~\cite{ClementeDRNT07} proposed building local maps independently and
stitching them with a hierarchical approach.

Ni et al. \cite{NiSD07} proposed an out-of-core bundle adjustment (BA) for large-scale
SfM. This method decomposes the data into multiple submaps, each of which has its own local
coordinate system for optimization in parallel. For global optimization, an out-of-core
implementation is adopted. Snavely~\etal~\cite{Snavely2008} proposed speeding up
reconstruction by selecting a skeletal image set for SfM and then adding other images with pose
estimation. Similarly, Konolige and Agrawal~\cite{KonoligeA08} selected a skeletal frame
set and used reduced relative constraints for closing large loops. Each skeleton frame
can actually be considered as a submap. A similar scheme is applied to iconic
views~\cite{LiWZLF08}, which are generated by clustering images with similar gist
features~\cite{OlivaT01}. In our work, a segment-based scheme is adopted, which first
estimates SfM for each sequence independently, and then aligns the recovered submaps.
Depending on estimation errors, we split each sequence to multiple segments, and perform
segment-based refinement. This strategy can effectively handle large data and quickly reduce
estimation errors during optimization.

Another line of research is to improve large-scale BA, which is a core component of SfM.
Agarwal~\etal~\cite{AgarwalSSS10} pointed out that connectivity graphs of Internet
image collections are generally much less structured and accordingly presented an inexact Newton type BA
algorithm. To speed up large-scale BA, Wu~\etal~\cite{WuACS11} utilized multi-core CPUs
or GPUs, and presented a parallel inexact Newton BA algorithm. Wu~\etal~\cite{Wu2013}
also proposed preemptive feature matching that reduces matching image pairs, and an
incremental SfM for full BA when the model is large enough. Pose graph
optimization~\cite{OlsonLT06,StrasdatMD10,KummerleGSKB11} was also widely used in
realtime SfM and SLAM~\cite{engel2014lsd,Mur-ArtalMT15}, which uses the relative-pose
constraints between cameras and is more efficient than full BA.

Most existing SfM approaches achieve reconstruction in an incremental way, which may risk
drifting or local minima when dealing with large-scale image sets.
Crandall~\etal~\cite{CrandallOSH11} proposed combining discrete and continuous
optimization to yield a better initialization for BA. By formulating SfM estimation as a labeling problem, belief propagation is employed
to estimate camera parameters and 3D points. In the
continuous step, Levenberg-Marquardt nonlinear optimization with additional constraints
is used. This method is restricted to urban scenes, and assumes that the vertical
vanishing point can be detected for rotation estimation, similar to the method proposed
by Sinha~\etal~\cite{SinhaSS10}. It also needs to leverage geotag contained in the
collected images and takes complex discrete optimization. In contrast, our segment-based
scheme can run on a common desktop PC with limited memory, even for large video data.

Real-time monocular SLAM methods~\cite{KleinM07,TanLDZB13,engel2014lsd,Mur-ArtalMT15}
typically perform tracking and mapping in parallel threads. The methods of
\cite{engel2014lsd,Mur-ArtalMT15} can close loops efficiently for large-scale scenes.
However, they could still have difficulty in directly handling multiple sequences, as
demonstrated in Figs.~\ref{fig:street-compare} and \ref{fig:garden-compare}.

\begin{table} [bpt]
\caption{Framework overview of ENFT-SFM}
\begin{tabular}{|ll|}
\hline
 1. &{\bf Consecutive point tracking} (Section~\ref{sec:consecutive}):\\
    &\begin{tabular}{lp{6.0cm}}1.1&
    Match the extracted SIFT features between consecutive frames with descriptor comparison.
 \end{tabular}     \\
    &\begin{tabular}{lp{6.0cm}}1.2& \hspace{-0.0in}Perform the second-pass matching to extend track lifetime.
 \end{tabular}     \\

 2.  &{\bf Non-consecutive track matching} (Section~\ref{sec:nonconsecutive}):\\
    &\begin{tabular}{lp{6.0cm}}2.1&\hspace{-0.0in}Use hierachical k-means to
    cluster the constructed invariant tracks.
 \end{tabular}     \\
 &\begin{tabular}{lp{6.0cm}}2.2&\hspace{-0.0in}Estimate
    the match matrix with the grouped tracks.
 \end{tabular}     \\
    &\begin{tabular}{lp{6.0cm}}2.3&\hspace{-0.0in}Detect overlapping subsequences
    and join the matched tracks.
 \end{tabular}     \\

 3.  &{\bf Segment-based coarse-to-fine SfM} (Section~\ref{sec:sfm}):\\
     &\begin{tabular}{lp{6.0cm}}3.1&\hspace{-0.0in}Estimate the submap for each sequence.
 \end{tabular}     \\
     &\begin{tabular}{lp{6.0cm}}3.2&\hspace{-0.0in}Match the common tracks among different sequences, and then use them to estimate the similarity transformations for each submap.
 \end{tabular}     \\
      &\begin{tabular}{lp{6.0cm}}3.3&\hspace{-0.0in}Use segment-based SfM
      to refine the aligned submaps.
 \end{tabular}     \\

 4.  &{\bf [Optional] Feature propagation with camera estimation}:\\
     &\begin{tabular}{lp{6.0cm}}4.1&\hspace{-0.0in}Quickly propagate features from sampled frames to others.
 \end{tabular}     \\
     &\begin{tabular}{lp{6.0cm}}4.2&\hspace{-0.0in}Quickly estimate camera poses for remaining frames.
 \end{tabular}     \\
\hline
\end{tabular}
\label{tab:overview}
\end{table}

\section{Our Approach}

Given a video sequence $V$ with $n$ frames, $V = \{I_t |t =
1,...,n\}$, our objective is to extract and match features in all
frames in order to form a set of \emph{feature tracks}. A feature
track $\mathcal{X}$ is defined as a series of feature points in
images: $\mathcal{X} = \{{\bf x}_t | t \in f(\mathcal {X})\}$, where
$f(\mathcal {X})$ denotes the frame set spanned by track $\mathcal
X$. Each SIFT feature ${\bf x}_t$ in frame $t$ is associated
with an appearance descriptor ${\bf p}({\bf x}_t)$ \cite{Lowe04} and
we denote all descriptors in a feature track as
$\mathcal{P_X}=\{{\bf p}({\bf x}_t) | t \in f(\mathcal {X})\}$.

With the detected $m$ features in all frames, finding matchable ones generally requires a
large amount of comparisons even using the k-d trees; meanwhile it inevitably induces
errors due to the fact that a large number of features make descriptor space hardly
distinctive, resulting in ambiguous matches. So it is neither reliable nor practical to
only compare the feature descriptors to form tracks. Our ENFT method has two main steps to
address this issue. The framework is outlined in Table~\ref{tab:overview}.

For reducing computation, we can extract one frame for every $3\sim5$ frames to
constitute a new sequence and then perform feature tracking on it. In the consecutive
tracking stage, we employ a two-pass matching strategy to extend the track lifetime. Then
in the non-consecutive tracking stage, we match common features in different
subsequences. With the obtained feature tracks, a segment-based SfM scheme is employed to
robustly recover the 3D structure and camera motion. Finally, if necessary, we propagate
feature points from sampled frames to others. Since the 3D positions of these features have
been computed, we can quickly estimate the camera poses of remaining frames with the
obtained 3D-2D correspondences.

\section{Consecutive Tracking}
\label{sec:consecutive}

For video sequences, feature tracks are typically obtained by matching features between
consecutive frames. However, due to illumination change, repeated texture, noise, and
large image distortion, features are easily dropped out or mismatched, resulting in
breaking many tracks into shorter ones. In this section, we propose an {\em improved
two-pass matching} strategy to alleviate this problem. The first-pass matching is the
same as our prior method~\cite{ZhangDJWB10}, which uses SIFT algorithm~\cite{Lowe04} with
RANSAC~\cite{Fischler81} to obtain high-confidence matches and remove outliers. In the
second pass matching, we firstly use the inlier matches to estimate a set of homographies
$\{H^k_{t,t+1}|k=1,...,N\}$ with multiple RANSAC
procedures~\cite{ZhangDJWB10,jin2008background}. To handle illumination change, we
estimate global illumination variation $L_{t,t+1}$ between images $I_t$ and $I_{t+1}$ by
computing the median intensity ratio between matched features. Here, $I_t$ denotes the gray scale image of frame $t$.

\begin{figure*}[bpt]
\centering
\includegraphics[width=0.85\linewidth]{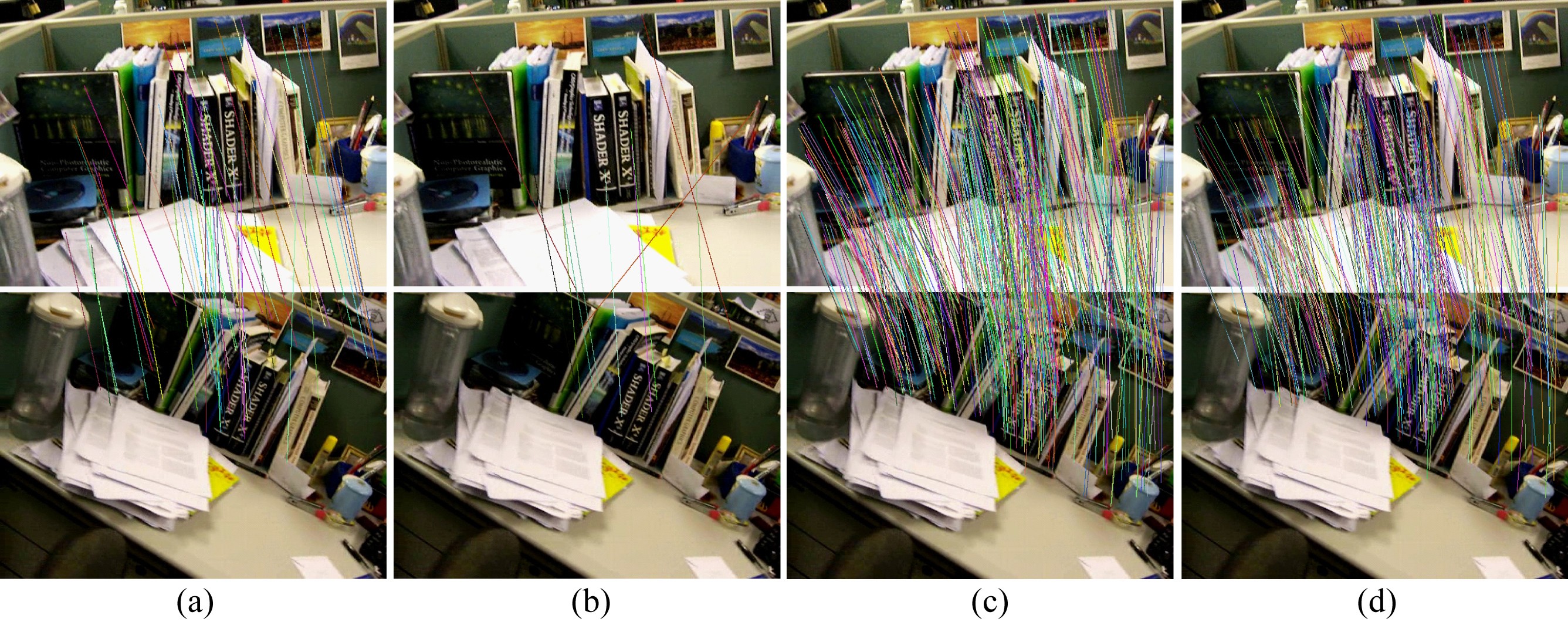}
\caption{Feature matching comparison. (a) First-pass matching by SIFT descriptor
comparison. There are $958$ features detected in the first image, but only 53 matches are
found. (b) Additional match by directly searching the correspondences along the epipolar
lines with SIFT descriptor comparison. Only 11 additional matches are found. (c)
Second-pass matching with outlier rejection. 399~(i.e.
$53+346$) matches are obtained. (d) Matching result of \cite{ZhangDJWB10}. 314 matches
are obtained.} \label{fig:match-compare}
\end{figure*}

\begin{figure}[tb]
\centering
\includegraphics[width=1.0\linewidth]{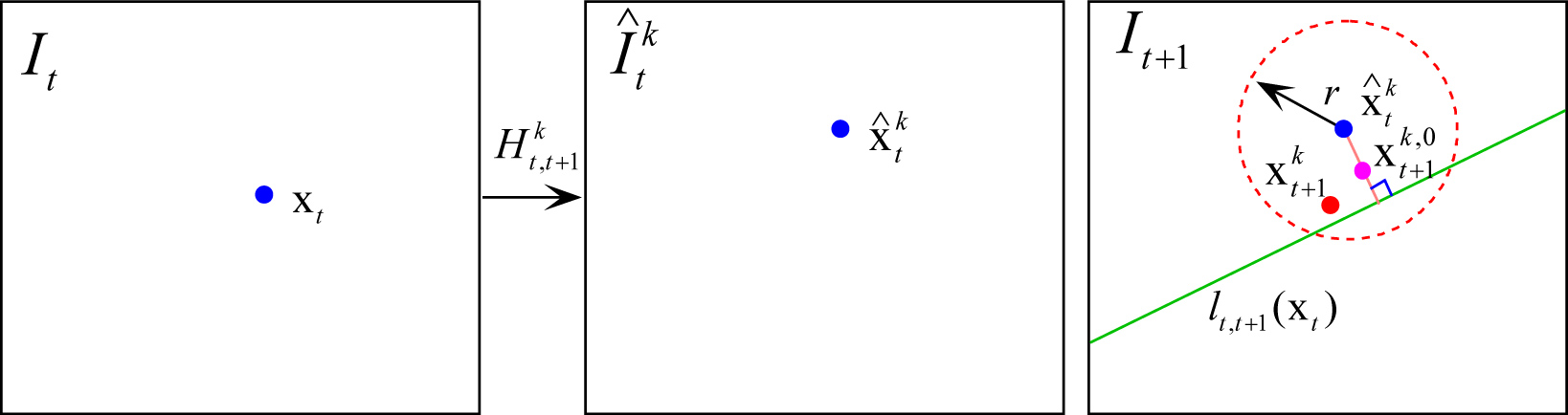}
\caption{Constrained spatial search with planar motion segmentation. Given homography
$H^k_{t,t+1}$, we rectify $I_t$ to $\hat I^k_t$ such that $\hat {\bf x}^k_t \sim
H^k_{t,t+1}{\bf x}_t$. Then we select the midpoint between $\hat {\bf x}^k_t$ and its
projection to $l_{t,t+1}({\bf x}_t)$ for initialization, and search the matched point by
minimizing (\ref{eq:epline-search}). The red dot ${\bf x}^k_{t+1}$ is the result.}
\label{fig:spatial-searching}
\end{figure}

We first linearly scale image $I_t$ with $L_{t,t+1}$, and then transform it with
homography $H^k_{t,t+1}$ to obtain the rectified image $\hat I^k_t$. Correspondingly,
${\bf x}_t$ in image $I_t$ is rectified to $\hat {\bf x}^k_t$ where $\hat {\bf x}^k_t
\sim H^k_{t,t+1}{\bf x}_t$ in $\hat I^k_t$. The distance between a 2D point ${\bf x}^k_{t
+ 1}$ and the epipolar line $l_{t,t+1}({\bf x}_t)$ is denoted by $d(\hat {\bf
x}^k_t,l_{t,t+1}({\bf x}_t))$. If $\hat {\bf x}^k_t$ largely deviates from the epipolar
line~(i.e., $d(\hat {\bf x}^k_t,l_{t,t+1}({\bf x}_t))
> \tau_e$), we reject $H^k_{t,t+1}$ since it does not
describe the motion of ${\bf x}_t$ well. For each remaining
$H^k_{t,t+1}$, we track ${\bf x}_t$ to ${\bf x}^k_{t + 1}$ by
minimizing the matching cost:
\begin{eqnarray}
S^k_{t,t+1}\hspace{-0.03in}({{\bf x}^k_{t + 1}}) \hspace{-0.03in}=\hspace{-0.1in} \sum\limits_{{\bf y} \in W} \hspace{-0.03in}{|{\hat I_t^k(\hat {\bf x}_t^k + {\bf y}) - {I_{t + 1}}({{\bf x}^k_{t + 1}} + {\bf y})} |^2} + \nonumber\\
\lambda_e d({\bf x}^k_{t + 1}, l_{t,t+1}({\bf x}_t))^2 +
\lambda_h ||\hat {\bf x}^k_t - {\bf x}^k_{t+1}||^2
\label{eq:epline-search}
\end{eqnarray}
where $\hat {\bf x}_t^k + {\bf y}$ are the points in the window $W$ centered at $\hat
{\bf x}_t^k$. Different from our prior method~\cite{ZhangDJWB10}, the matching cost
incorporates two geometric constraint terms, which encourage ${\bf x}^k_{t + 1}$ to be
along the epipolar line and obey homography. $||.||$ is the Euclidean distance and $|.|$
is the absolute value. The corresponding weights are $\lambda_e =
|W|\sigma_c^2/\sigma_e^2$ and $\lambda_h = |W|\sigma_c^2/\sigma_h^2$, where $\sigma_c$,
$\sigma_e$, and $\sigma_h$ account for the uncertainty of intensity, epipolar geometry
and homography transformation respectively. In our experiments, these values are by
default $\sigma_c = 0.1$~(for intensity values normalized to $[0, 1]$), $\sigma_e = 2$
and $\sigma_h = 10$. Note that $\sigma_h$ is relatively large because we do not require
the points to strictly lie on the same plane. As long as the point is near the plane,
$H_k$ can alleviate the major distortion and provide a better matching condition.

Similar to KLT tracking, we solve for $S_{t,t+1}({{\bf x}^k_{t + 1}})$ iteratively by
taking the partial derivative w.r.t. ${\bf x}^k_{t+1}$ and setting it to zero:
\begin{equation}
\frac{{\partial S^k_{t,t+1}({\bf x}^k_{t+1})}}{\partial{\bf x}^k_{t+1}} =
0. \label{eq:match-partial}
\end{equation}
${I_{t + 1}}({\bf x} + \Delta{\bf x})$ is approximated by a Taylor
expansion truncated up to its first order:
\begin{equation}
{I_{t + 1}}({\bf x} + \Delta{\bf x}) \approx {I_{t + 1}}({\bf x})
+ g_{t + 1}^\top({\bf x}) \cdot {\Delta {\bf x}}
\label{eq:color-taylor}
\end{equation}
where $g_{t + 1}^\top$ is the image gradient in the $(t+1)^{th}$
frame. With the computed gradients, we propose an iterative solver
to optimize (\ref{eq:epline-search}) by first initializing ${\bf
x}_{t+1}$ as the midpoint between $\hat {\bf x}^k_t$ and its
projection to $l_{t,t+1}({\bf x}_t)$, as shown in Fig.
\ref{fig:spatial-searching}. Then we iteratively update ${\bf
x}_{t+1}$ by solving (\ref{eq:match-partial}). In iteration $i+1$,
${\bf x}_{t+1}$ is updated as
\[
{\bf x}^{k,(i+1)}_{t+1} = {\bf x}^{k,(i)}_{t+1} + \Delta{\bf x}
\]
where ${\bf x}^{k,(i)}_{t+1}$ denotes the value of ${\bf x}^k_{t+1}$
in iteration $i$. This procedure continues until $\Delta{\bf x}$ is
sufficiently small.

\if 0
\begin{algorithm}[bpt]
\caption{Planar Motion Segmentation} \label{alg:homography}
1. Put all matches into a set $\Omega$.\\
2. For $k=1,....,N_{\max}$, ~~~\%$N_{\max}$ is the maximum number
of the
homographies.\\
\begin{tabular}{lp{7cm}}~~~&
2.1 ~Use RANSAC to estimate homography $H^k_{t,t+1}$ that has the
maximum inliers. \\
&2.2 ~Remove the inliers from $\Omega$. If the size of $\Omega$ is
small enough, stop; otherwise, continue.
\end{tabular}
\end{algorithm}
\fi

The found match is denoted as ${\bf x}^k_{t+1}$. With the set of homographies
$\{H^k_{t,t+1}|k=1,...,N\}$, we can find several matches $\{{\bf x}^k_{t+1}|k=1,...,N\}$.
Only the best one $j = \mathop {\min_{k} }\limits {\sum_{{\bf y} \in W} {|{\hat
I_t^k(\hat {\bf x}_t^k + {\bf y}) - {I_{t + 1}}({{\bf x}^k_{t + 1}} + {\bf y})} |}}$ is
kept.

In case the feature motion cannot be described by any homographies or feature
correspondence is indeed missing, the found match is actually an outlier. We detect it
with the following conditions:
\begin{displaymath}
\left\{ \begin{array}{l}
\sum_{{\bf y} \in W} {|{\hat I_t^j(\hat {\bf x}_t^j + {\bf y}) -
{I_{t + 1}}({{\bf x}^j_{t + 1}} + {\bf y})}|} >
\tau_c |W|;\\
d({\bf x}^j_{t+1},l_{t,t+1}({\bf x}_t)) > \tau_e;\\
||\hat {\bf x}^j_t - {\bf x}^j_{t+1}||> \tau_h.
\end{array}
\right.
\end{displaymath}
These conditions represent the constraints of color constancy, epipolar geometry and
homography respectively. If any of them is satisfied, ${\bf x}^j_{t+1}$ is treated as an
outlier. $\tau_c$ is set to a small value~(generally $0.02$ in our experiments) since the image is
rectified. The remaining two parameters are $\tau_e = 2$ and $\tau_h = 10$. Considering
points may not strictly undergo planar transformation, $\tau_h$ is set to a relatively
large value.

Fig.~\ref{fig:match-compare}(c) shows the result after the second-pass matching. Compared
to our prior method~\cite{ZhangDJWB10} (Fig.~\ref{fig:match-compare}(d)), the improved
two-pass matching method does not need to perform additional KLT matching. It thus runs
faster. The computation time is only $18$ms with GPU acceleration on a NVIDIA GTX780
display card. The number of credible matches also increases.

The two-pass matching can produce many long tracks. Each track has a group of
descriptors. They are similar to each other in the same track due to the matching
criteria. We compute average of the descriptors over the track, and denote it as {\it
track descriptor} ${\bf p}(\mathcal X)$. It is used in the following non-consecutive
track matching.

\section{Non-Consecutive Track Matching}
\label{sec:nonconsecutive}

In this stage, we match features distributed in different subsequences, which is vital
for drift-free SfM estimation. If we select all image pairs in a brute-force manner, the
process can be intolerably costly for a long sequence. A better strategy is to estimate
content similarity among different images first. We propose a non-consecutive track
matching (NCTM) method to address this problem.

There are two steps. In the first step, similarity of different images is coarsely
estimated by constructing a $n \times n$ symmetric match matrix $M$, where $n$ is the
number of frames. $M(i, j)$ stores overlapping confidence between images $I_i$ and $I_j$.
We use the same method of \cite{ZhangDJWB10} to quickly estimate the initial matching
matrix $M$, which first uses hierarchical K-means to cluster the track descriptors and
then compute the similarity confidence of frame pairs by counting the number of
potentially matched tracks that are clustered into the same leaf node.

\begin{figure}[tb!]
\centering
\includegraphics[width=1.0\linewidth]{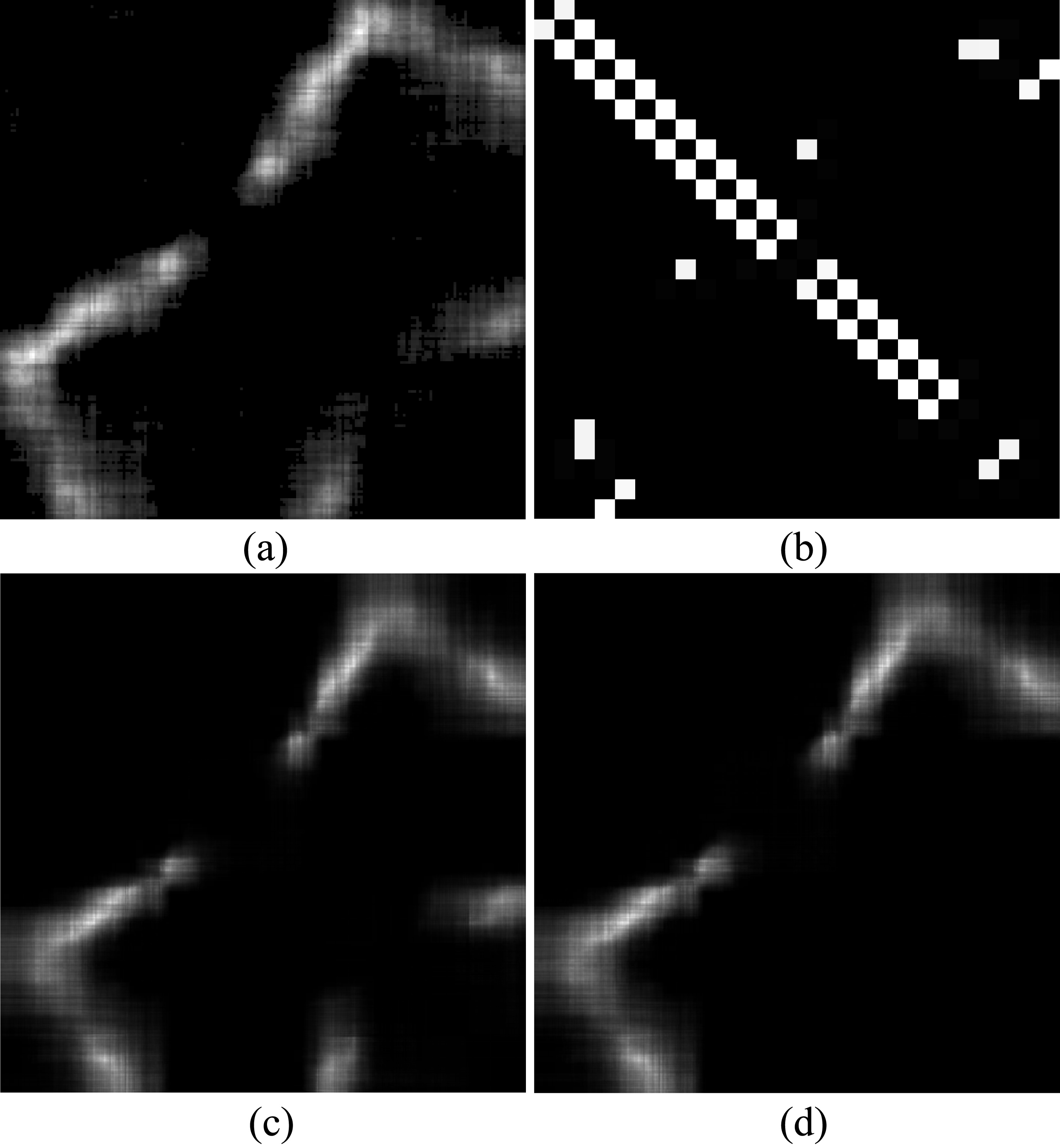}
\caption{Match matrix estimation for the ``Desktop" sequence containing 941 frames. (a)
Our initially estimated match matrix based on the keyframes. The matrix size is scaled
for visualization. (b) Estimated match matrix by FAB-MAP~\cite{CumminsN11} on the
re-sampled sequence that contains 26 frames. (c) The final match matrix for all frames
after our non-consecutive matching based on (a). (d) The final match matrix after our
non-consecutive matching based on (b). } \label{fig:desktop}
\end{figure}

For acceleration, we only select long tracks that span 5 or more keyframes to estimate
overlap confidence. In our experiments, for the ``Desktop" sequence, the initial match
matrix estimation only takes 1.08 seconds, with a total of $5,935$ selected feature
tracks. Fig.~\ref{fig:desktop}(a) shows the initially estimated match matrix for the
``Desktop'' sequence. Bright pixels are with high overlapping confidence where many
common features exist. Because we exclude track self-matching, the diagonal band of estimated match matrix has no value. Our method handles dense image sequences, unlike
FAB-MAP~\cite{CumminsN08,CumminsN11} that assumes sparsely sampled ones. When applying FAB-MAP to the
original ``Desktop" sequence, no loop is detected. So we manually sample the original
sequence until common points between adjacent sampled frames are no more than 100. This
generates 26 sampled frames. As shown in Fig.~\ref{fig:desktop}(b), in this case, a few
overlapping image pairs are identified by FAB-MAP; but they are not enough to match many
common features.

In the second step, with the initially estimated match matrix, we select the frame pairs
with maximum overlapping confidence to perform feature matching, and update the match
matrix iteratively. Matrix estimation and non-consecutive feature matching are benefitted
from each other to simplify computation. Fig.~\ref{fig:desktop}(c) shows our finally
estimated match matrix.

For speedup, we extract keyframes based on the result of consecutive feature tracking
described in Section~\ref{sec:consecutive}. Frame $1$ is selected as the first keyframe.
Then we select frame $i$ as the second keyframe if it satisfies $N_1(1, i) \geq m_1$ and
$N_1(1, i+1) < m_1$, where $N_1(i, j)$ denotes the number of common features between
frames $i$ and $j$. Other keyframes are selected as follows. For the two recent keyframes
with indices $i_1$ and $i_2$ in the original sequence, we select frame $j$~($j>i_2$) as
the new keyframe if it is the farthest one from $i_2$ that satisfies $\{N_1(i_1, j) \geq
m_1$, $N_2(i_1, i_2, j) \geq m_2\}$, where $N_2(i_1, i_2, j)$ denotes the number of
common points among the three frames $(i_1, i_2, k)$. This step is repeated until all
frames are processed. In our experiments, $m_1 = 100\sim500$ and $m_2 = 50\sim300$.
Without special notice, the following procedures are only performed on keyframes.

%\subsection{Fast Match Matrix Estimation}\label{sec:matching-mtx}

\subsection{Non-Consecutive Track Matching}
\label{sec:nonconsecutive-matching}

Since the number of common features between two frames can be coarsely reflected by the
initially estimated match matrix $M$, we select a frame pair $(t_1^0, t_2^0)$ with the
largest value in $M$ to start matching. After matching $(t_1^0, t_2^0)$, the set of
matched track pairs $C_{\mathcal X} = \left\{ ({\mathcal X}_1, {\mathcal X}_2) \right\}$
approximately represent the number of common tracks for neighboring frame pairs. The
matched track pairs in frame pair $(t_1, t_2)$ can be expressed as
\begin{equation}
\begin{split}
C_{\mathcal X}(t_1,t_2) = \{
({\mathcal X}_1,{\mathcal X}_2) | &t_1 \in f({\mathcal X}_1), t_2 \in f({\mathcal X}_2),\\
                                  &({\mathcal X}_1,{\mathcal X}_2) \in C_{\mathcal X}
\}.
\end{split}
\label{eq:Cx}
\end{equation}
The number of common features in $(t_1, t_2)$ can be approximated by $|C_{\mathcal
X}(t_1,t_2)|$ as long as $(t_1, t_2)$ shares sufficient common tracks with $(t_1^0,
t_2^0)$. We maintain an updating match matrix $M^{\ast}$, computed as
\begin{equation}
M^{\ast}(t_1,t_2) = |C_{\mathcal X}(t_1,t_2)| \label{eq:Mast}
\end{equation}
to propagate
the overlapping confidence from $(t_1^0, t_2^0)$ toward the
neighboring frame pairs, and determine where the next matching
should be performed. Details are given below.

\vspace{0.1in} \noindent {\bf Main Procedure:} We first detect the largest element
$(t_1^0, t_2^0)$ in $M$. The value of $M(t_1^0, t_2^0)$ is also denoted as $M_{\max}$. If
$M(t_1^0, t_2^0)$ is larger than a threshold, several common features may exist. After
matching $(t_1^0, t_2^0)$, we collect and put the matched track pairs into $C_{\mathcal
X}$ and update $M^{\ast}$ according to Eq. (\ref{eq:Mast}). In particular, we set
$M^{\ast}(t_1^0, t_2^0) = 0$, indicating $(t_1^0, t_2^0)$ is matched. Next, we repeatedly
select the largest element $(t_1^k, t_2^k)$ in the {updating} matrix $M^{\ast}$, match
$(t_1^k, t_2^k)$, and update $C_{\mathcal X}$ and $M^{\ast}$ accordingly. This procedure
continues until $M^{\ast}(t_1^k, t_2^k) < 50$. Then we go to another region by
re-detecting the brightest point in $M$ that has not been processed. The step ends if the
brightest value is smaller than $0.1M_{\max}$.

\begin{table} [tb]
    {
\caption{Non-consecutive track matching comparison between the method of
\cite{ZhangDJWB10} and ours for the ``Desktop'' and ``Circle" sequences.}
        \footnotesize
        \begin{center}
            \begin{tabular}{|c|c|c|c|c|}
\hline  Methods & \multicolumn{2}{c|}{Desktop Sequence} & \multicolumn{2}{c|}{Circle Sequence}\\
                &  Merged Tracks & Time & Merged Tracks & Time  \\
                \hline  \cite{ZhangDJWB10} & 16, 279 & 81s & 101, 948 & 132s \\
                \hline  Our method & 16, 827 & 35s & 102, 583 & 55s \\
                \hline
            \end{tabular}
        \end{center}

        \label{tab:track-matching-statistic}}
\end{table}

\vspace{0.1in} \noindent {\bf Frame pair matching and outlier rejection:}
When entering a new bright region, we
perform the classical 2NN matching for $(t_1^0, t_2^0)$. Then each matching pair $(t_1^k,
t_2^k)$ is detected from the {updating} matrix $M^{\ast}$. Thus there are
$M^{\ast}(t_1^k, t_2^k)$ common features found previously. We use these matches to
estimate the fundamental matrix $F_{t^k_1,t^k_2}$ of frame pair $(t_1^k, t_2^k)$, and
re-match those outlying features along the epipolar lines. We further search the
correspondences for other unmatched features along epipolar lines.

Along with the fundamental matrix estimation between $t_1^k$ and $t_2^k$, these
$M^{\ast}(t_1^k, t_2^k)$ matches are classified into inliers and outliers. Since only
part of matches are used to estimate $F_{t^k_1,t^k_2}$, the estimated $F_{t^k_1,t^k_2}$
could be biased. So we do not reject outliers immediately. Fortunately, each matched
track pair $(\mathcal{X}_1, \mathcal{X}_2)$ undergoes multi-pass epipolar verification
during processing the whole bright region. We record all the verification results for
each $(\mathcal{X}_1, \mathcal{X}_2)$, and determine inliers/outliers after all bright
regions are processed. Suppose $(\mathcal{X}_1, \mathcal{X}_2)$ is classified as an
inlier match $N_I$ times and as an outlier match $N_O$ times. We reject $(\mathcal{X}_1,
\mathcal{X}_2)$ if $N_I < s \cdot N_O$~($s=1\sim4$ in our experiments). In addition, we
use the following strategy to remove the potential matching ambiguity. For example, a
track ${\mathcal X}_1$ may find two corresponding tracks ${\mathcal X}_2$ and ${\mathcal
X}'_2$, where ${\mathcal X}_2$ and ${\mathcal X}'_2$ have overlapping frames. So the
track matches $({\mathcal X}_1, {\mathcal X}_2)$ and $({\mathcal X}_1, {\mathcal X}'_2)$
conflict with each other. In this case, we simply select the best match with the largest
$N_I$, and regard the other as an outlier.

\vspace{0.1in} \noindent {\bf Benefits:} The proposed matching method outperforms
previous ones in the following aspects. In our prior method~\cite{ZhangDJWB10}, a
rectangular region in the roughly estimated match matrix $M$ is sought each time and
local exhaustive track matching is performed for all frame pairs in it. It could involve
a lot of unnecessary matching for non-overlapping frames and repeated feature comparison.
Our current scheme only selects the frame pairs with sufficient overlapping, and matches
each pair of frames and most tracks at most once. As shown in
Table~\ref{tab:track-matching-statistic}, compared to the method of \cite{ZhangDJWB10},
our new non-consecutive track matching algorithm is more efficient. Both methods are
implemented without GPU acceleration.

Standard image matching is to find a set of most similar images given the query one. This
scheme has been extensively used in large-scale SfM~\cite{AgarwalSSSS09,FrahmGGJRWJDCL10}
and realtime SLAM systems for loop closure
detection~\cite{CastleKM08,ClippLFP10,LimFP11}. It, however, also may involve unnecessary
matching for unrelated frame pairs and miss those with considerable common features. It
is because image similarity based on appearance may not be sufficiently reliable. In
contrast, we progressively expand frames with track matching. The expansion is not fully
related to the initial match matrix. Therefore a very rough matrix is enough to give a good
starting point. Practically, as long as there is one good position, our system can extend
it to the whole overlapping region accurately. To verify this, we provided two refined
match matrices based on two different rough match matrices, as shown in
Figs.~\ref{fig:desktop}(a) and (b). Although the two initially estimated match matrices
are different and only based on keyframes, the finally estimated match matrices after our
non-consecutive track matching are quite similar~(except the bottom right area, where the
initial match matrix by FAB-MAP does not provide any high confidence elements), which
demonstrates the effectiveness of our method.

\subsection{SfM for Multiple Sequences}
\label{sec:multi-video-tracking} Our method can be naturally extended to handle multiple
sequences. Given one or multiple sequences, we first split long ones, making each new
sequence generally contains only $1000\sim3001$ frames. The splitted neighboring sequences
can contain some overlapping frames for reliable matching. The sequence set is denoted as
$\{V_i|i=1,...,n\}$. Then we apply our feature tracking to each $V_i$, and estimate its 3D
structure and camera motion using a keyframe-based incremental SfM scheme similar to that of \cite{ZhangQHWHB07}.
The major modification is that we use known intrinsic camera parameters, and simply
select an initial frame pair that has sufficient matches and a large baseline to start
SfM. For each sequence pair, we use the fast matching matrix estimation algorithm
\cite{ZhangDJWB10} to estimate the rough match matrix such that related frames in any two
different sequences can be found and common features can be matched by the algorithm
introduced in Section~\ref{sec:nonconsecutive-matching}. Then we use the segment-based
SfM method described in Section~\ref{sec:sfm} to efficiently recover and globally
register 3D points and camera trajectories, as shown in Fig.~\ref{fig:garden}(b).

\section{Segment-based Coarse-to-Fine SfM}
\label{sec:sfm}

\begin{figure}[tb]
\centering
\includegraphics[width=1.0\linewidth]{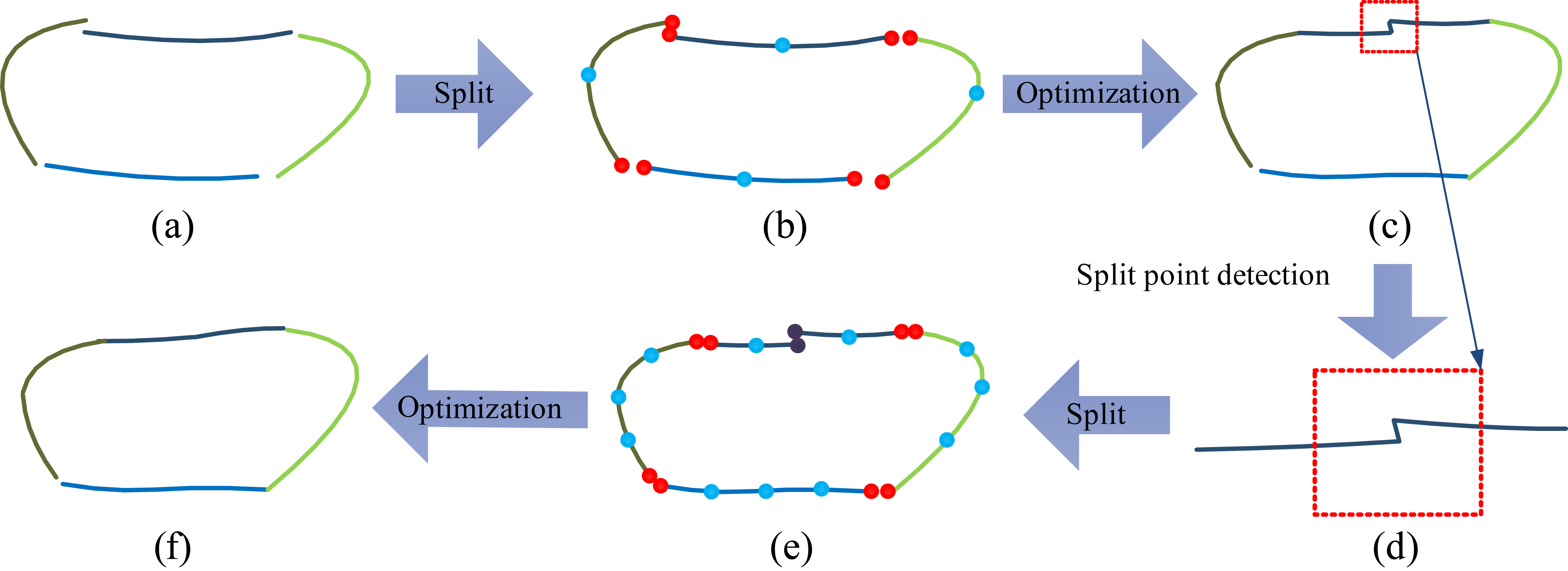}
\caption{Segment-based coarse-to-fine refinement. (a) Recovered camera trajectories marked
with different colors. (b) Each sequence is split into 2 segments where endpoints and
split points are highlighted. (c) Refined camera trajectories after the first iteration, where
errors are redistributed. (d) ``Split points", which are joints of largely inconsistent
camera motion for consecutive frames. (e) Sequence separation by split points. Two dark points denote the splitted two consecutive frames in a split point. (f)
Refined camera trajectories after 2 iterations. } \label{fig:coarse-to-fine}
\end{figure}

With the independently reconstructed sequences and matched common tracks, we align them
in a unified 3D coordinate system. For a long loopback sequence, error accumulation could
be serious, making traditional bundle adjustment stuck in local optimum. It is because the
first a few iterations of bundle adjustment aggregate accumulation errors at the joint
loop points, which are hard to be propagated to the whole sequence. To address this
problem, we split each sequence into multiple segments -- each is with a similarity
transformation. Only these transformations and overlapping points across different
segments are optimized. We name it segment-based bundle adjustment and illustrate it in
Fig.~\ref{fig:coarse-to-fine}. Lim~\etal~\cite{LimFP11} performed global adjustment by
clustering keyframes into multiple disjoint sets~(i.e. segments), which is conceptually similar to our idea. But the geodesic-distance-based segmentation to cluster frames could make inconsistent
structure be put into a single body, complicating alignment-error reduction. This method also did not adaptively split the segments in a coarse-to-fine way to minimize the accumulation error. Local
optimization within each body may not sufficiently minimize the error which is mainly caused by global
misalignment.

In the beginning, we order all sequences and define the one that
contains the maximum number of tracks merged with others as the
reference. Without losing generality, we define it as sequence
$\#1$, denoted as $V_1$. Its local 3D coordinate system is also
set as the reference. Then with the common tracks among different
sequences, we can estimate the coordinate transformation for each
sequence $j$ (i.e., $V_j$), denoted as $T_j=(s_j, R_j,t_j)$, where
$s_j$ is the scale factor, $R_j$ is the rotational matrix, and
$t_j$ is the translation vector. For the reference sequence, $s_1$
have value 1, $R_1$ is an identity $3\times3$ matrix, and $t_1 =
(0,0,0)^\top$.

\begin{figure*}[tb]
\centering
\includegraphics[width=1.0\linewidth]{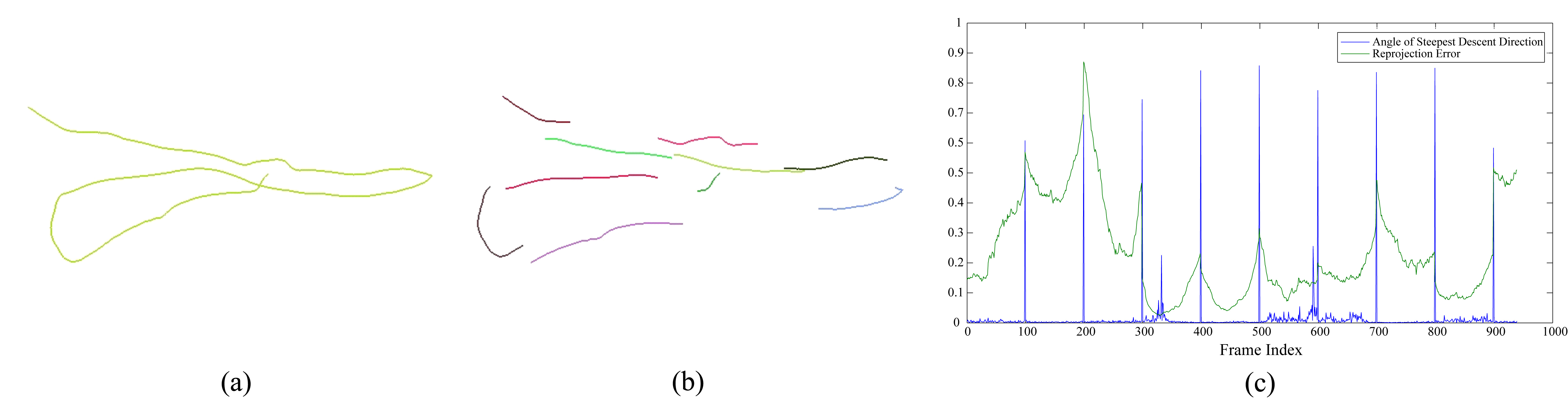}
\caption{Split point detection. (a) Original camera trajectory of the ``Desktop''
sequence. (b) Splitted camera trajectories. Each segment contains 100 frames. (c)
Re-projection errors~(green curve) and angles of steepest descending direction~(blue
curve). Values are all normalized to [0,1] for better comparison. The angle more accurately reflects the split
result quality compared to the re-projection error.} \label{fig:split}
\end{figure*}

Each segment is assigned with a similarity transformation, and the relative camera motion
between frames in each segment is fixed, so that the number of variables is small enough
for efficient optimization. Different from \cite{LimFP11}, which clusters frames using
geodesic distances, we propose clustering neighboring and geometrically consistent frames
into segments. The position at which two consecutive frames are inconsistent is defined
as a ``split point". We project the common points in each consecutive frame pair into the
two images and check the re-projection error.

However, directly detecting the split points according to reprojection error is not
optimal since it is generally large at loop closure points. Splitting such frame pairs
does not help. We instead find split points that the re-projection error is most likely
to be reduced. Assume each frame $k$ is associated with a small similarity transformation
$T_k$, which is parameterized as a $7$-vector $a_k$~(three Rodrigues components for
rotation, 3D translation and scale). If we minimize the re-projection error w.r.t. $a_k$,
the steepest descent direction is
\begin{equation}
g_k = \sum_{i=1 \cdots N_k} {A_i^Te_i}
\end{equation}
where $N_k$ is the number of points visible in frame $k$, and
$A_i$ is the Jacobian matrix $A_i =
\partial{\pi(P_kX_i)}/\partial{a_k}$. $\pi$ is the projection
function. $e_i$ is the re-projection error $e_i = {\bf x}_i - \pi(P_kX_i)$, which is
reduced along the direction of $g_k$. For two consecutive frames $(k,k+1)$, if their
$g_k$ and $g_{k+1}$ have similar directions, their re-projection errors both can be
reduced with the same similarity transformation. Otherwise, these two frames are better
to be assigned to different segments. The inconsistency between two consecutive frames is
defined as the angle between the two steepest descent directions
\begin{equation}
C(k, k+1) = \arccos\frac{g_k^T \cdot g_{k+1}}{||g_k|| \cdot ||g_{k+1}||}.
\end{equation}
For verification, we group every 100 consecutive frames into one segment for the
``Desktop"~(Fig.~\ref{fig:split}(a)), and apply a certain transformation to each segment
(Fig.~\ref{fig:split}(b)). As expected, the re-projection errors distribute in the whole
overlapping regions. In contrast, the angle between the steepest descent directions
reliably reflects the splitting result.

We progressively segment the sequences. At the $t^{th}$ iteration, each sequence is
divided into $2^t$ segments. We compute $C(k,k+1)$ for all $k$ and detect the $2^t-1$
split points with the largest $C(k,k+1)$. In order to evenly spread the split points across the whole sequence, we perform non-maximal suppression during selecting split points. 
While selecting the largest one, its neighboring $\frac{N_j}{2^t}$ candidates~($N_j$ is the number of frames in sequence $j$) are suppressed and then select the next largest one from the remaining ones with non-maximal suppression. This procedure is repeated until $2^t-1$ split points are selected. 
We put the consecutive frames in between two adjacent split points into a segment, and use the method described as follows to estimate
the similarity transformations and submaps jointly for all segments. When the
optimization is done, we detect split points for each sequence again, and re-separate the
sequence into multiple segments. We can repeat this process until the average reprojection
error is below a threshold or each segment contains only one frame. Errors are progressively
propagated and reduced. The procedure of our segment-based coarse-to-fine refinement
scheme is illustrated in Fig.~\ref{fig:coarse-to-fine}.

\begin{table*}
	[tb] {
		\caption{Running time of ENFT-SFM.}
		\footnotesize
		\begin{center}
			\begin{tabular}{|@{\hspace{1mm}}c@{\hspace{1mm}}|@{\hspace{1mm}}c@{\hspace{1mm}}|@{\hspace{1mm}}c@{\hspace{1mm}}|@{\hspace{1mm}}c@{\hspace{1mm}}|@{\hspace{1mm}}c@{\hspace{1mm}}|@{\hspace{1mm}}c@{\hspace{1mm}}|@{\hspace{1mm}}c@{\hspace{1mm}}|c|@{\hspace{1mm}}c@{\hspace{1mm}}|@{\hspace{1mm}}c@{\hspace{1mm}}|c|@{\hspace{1mm}}c@{\hspace{1mm}}|c|}
				\hline   &  & & & & \multicolumn{2}{c|}{Feature Tracking} & \multicolumn{3}{c|}{SfM Estimation} & \multicolumn{2}{c|}{Propagation} & \\
				Datasets & Frames &  Step & Sampled & Resolution & CPT & KNCTM & Submap & Align & Refine & Feature & Camera & Reprojection\\
				& &  & Frames & &  & & Estimation &  & & Propagation & Estimation & Error\\
				\hline  Desktop &  $941$ & 1 & 941 & $640\times480$ & 46.5s & 5.8s & 14.1s & - & - & - & - & 1.26 pixels\\
				\hline  Circle &  $2,129$ & 3 & 710 & $960\times540$ & 63.1s & 40.9s & 13.8s & - & - & 13.9s & 4.2s & 1.07 pixels\\
				\hline  Street &  $22,799$ & 5 & 5,537 & $960\times540$ & 7.4 min. & 7.5 min. & 176.0s & 3.6s & 32.0s & 3.4 min. & 65.1s & 2.49 pixels\\
				\hline  Garden &  $95,476$ & 3 or 5 & 21,791 & $960\times540$ & 27.4 min. & 31.2 min. & 588.1s & 2.5s & 130.4s & 15.6 min. & 3.7 min. & 2.28 pixels\\
				\hline
			\end{tabular}
		\end{center}
		\label{tab:time-statistics}}
\end{table*}

\vspace{0.1in} \noindent {\bf Algorithm Details:~} Suppose the
number of detected split points among all $n$ sequences is $m$. We
break the sequences into a total of $n' = n+m$ segments. Each
of them is with a similarity transformation
$T^w_j=(s_j^w,R_j^w,t^w_j)$, where $j=1, \cdots, n'$, w.r.t. the
world coordinate. We use BA to refine the
reconstructed 3D feature points with these
similarity transformations. Different from traditional BA, the
camera parameters inside each segment are fixed, we thus only
update the similarity transformation. The procedure is to first
transform one 3D point in the world coordinate to a local one with
parameters $T^w$. Then traditional perspective camera projection
is employed to compute the re-projection error. Our BA function is
written as

{\small
\begin{eqnarray}
\mathop {\min}\sum\limits_{i = 1}^{N'} \sum\limits_{j = 1}^{n'}
\sum\limits_{k = 1}^{n_j} w_{i,j,k}\| \pi
(K_{j,k}(R_{j,k}(s_j^wR_j^w{X_i} + t_j^w)+t_{j,k}))\nonumber\\
- {{\bf x}_{i,j,k}} \|^2 \label{eq:BA}
\end{eqnarray}
}where $n_j$ is the number of frames in the $j$-th segment,
$N'$ is the number of the 3D feature points, and $n'$ is the
number of the segments. $\pi$ is the projection function. ${\bf
x}_{i,j,k}$ is the image location of $X_i$ in the $k$-th frame of
the $j$-th subsequence. $K_{j,k}$, $R_{j,k}$, and $t_{j,k}$ are
the intrinsic matrix, rotational matrix, and translation vector,
respectively. $w_{i,j,k}$ is defined as
{\small
\[
w_{i,j,k} = \left\{ \begin{array}{l}
1,~~~{\rm If~point}~ i~ {\rm{is~ visible~ in~ frame~}} k~ {\rm{in~ sequence}}~ j\\
0,~~~{\rm Otherwise}
\end{array} \right.
\]}
We use Schur complement and forward substitution~\cite{Triggs99} to solve the normal
equation, which separates the updating of rigid transformation and of 3D points in each
iteration. It reduces the large linear system to a linear symmetric one with scale
$7n'\times 7n'$ for updating transformation. It makes 3D point estimation much cheaper
because each point can be updated independently by solving a $3\times3$ linear symmetric
system. Moreover, since only a few segment pairs share common points, the Schur
complement is rather sparse.
In SBA~\cite{lour09}, the system of Schur complement was explicitly
constructed and solved by Cholesky decomposition. Wu~\etal~\cite{WuACS11} implicitly
built the Schur complement for parallel computing. They did not take full advantage of
the sparsity property. For acceleration, sSBA ~\cite{Konolige10} proposed to utilize the sparse structure of Schur complement and solve it with sparse Cholesky decomposition. We also utilize the sparsity and solve it with efficient preconditioned conjugate gradient similar to that of \cite{WuACS11}, which can significantly reduce the computation.

Because the size of the linear system is actually determined by $n'$, we can estimate $n'$
based on the available memory. Once the size $n'$ linear system is reached, SfM
refinement is performed in the following two steps. In the first step, we only select the
$m=n'-n$ split points to split the sequences, and solve (\ref{eq:BA}) to refine the result. In the second step, we perform a local BA for each sequence $j$ iteratively by re-splitting sequence $j$ to multiple
segments with detected split points and refining them by solving (\ref{eq:BA}) while
fixing cameras and 3D points in other sequences. This process stops when all sequences
are processed. This strategy makes it possible to efficiently and robustly handle large
data with limited memory consumption.

Finally, we fix the 3D points and estimate the camera poses respectively for all frames.
During the course of iterations, errors are quickly reduced.

\section{Experimental Results}

\begin{table}
	[tb] {\small
		\caption{Performance of different algorithms.}
		\begin{center}
			\begin{tabular}{|c|c|c|}
				\hline  Algorithms & Running Time & Average Track Length\\
				\hline  C-SIFT & 38.4s & 1.73 \\
				\hline  CPT & 63.1s & 2.28  \\
				\hline  CPT+NCTM & 150.7s & 3.10 \\
				\hline  CPT+KNCTM & 104.0s & 2.68 \\
				\hline  BF-SIFT & 1086.4s & 2.71 \\
				\hline
			\end{tabular}
		\end{center}
		\label{tab:statistics}}
\end{table}

We evaluate our method with several challenging sequences. Running time is listed in
Table~\ref{tab:time-statistics} excluding I/O, which is obtained on a desktop PC with an
Intel i7-4770K CPU, 8GB memory, and a NVIDIA GTX780 graphics card. The operating system
is 64-bit Windows 7. Only the feature tracking component is accelerated by
GPU. We use 64D descriptors for SIFT features. Our SIFT GPU implementation is
inspired by \cite{siftgpu07wu} but runs faster. For SfM estimation, we optimize the
code by applying SSE instructions, but only use a {\bf single thread} without parallel
computing. For the sequences captured by us, since the intrinsic matrix is known, we
optimize the SfM code by incorporating this prior to improve the robustness and
efficiency. Garden dataset contains 6 sequences, which are further splitted into 37
shorter sequences, from which we sample the frames by setting the step to 3 or 5. The source code and datasets can be found in our project website\footnote{http://www.zjucvg.net/ls-acts/ls-acts.html}.

As our consecutive point tracking can handle wide-baseline images, frame-by-frame
tracking is generally unnecessary. For our datasets listed in Table~\ref{tab:time-statistics}, we usually extract one
frame for every $3\sim5$ frames to apply feature tracking. We quickly propagate the
feature points to other frames by KLT with GPU acceleration. This trick further saves
computation. In addition, in order to reduce image noise and blur, for each input frame
$I_t$, we perform matching with two past frames. One is the last frame $I_{t-1}$, and the
other (denoted as $I_{t'}$) is the farthest frame that shares over 300 common features
with $I_{t-1}$. Note that only a small number of features in $I_{t'}$ need to be matched
with $I_t$, which does not increase computation much.

\subsection{Quantitative Evaluation of Feature Tracking}
We compare the feature tracking methods of consecutive SIFT
matching~(C-SIFT), our consecutive point tracking~(CPT),
brute-force SIFT matching~(BF-SIFT), our consecutive point tracking with
non-consecutive track matching~(CPT+NCTM), our consecutive point tracking with keyframe-based
non-consecutive track matching~(CPT+KNCTM).

C-SIFT extracts and matches SIFT features only in consecutive frames. It is a common
strategy for feature tracking. The advantage is that the complexity is linear to the
number of frames. However, feature dropout could occur due to global indistinctiveness or
image noise, which causes producing many short tracks. The brute-force SIFT matching
exhaustively compares extracted SIFT features, whose complexity is quadratic to the
number of processed frames. In comparison, the complexity of our method~(CPT+NCTM) is
linear to the number of processed frames and the number of overlapping frame pairs while high quality results are guaranteed.

\begin{figure*}[t!]
	\centering
	\includegraphics[width=1.0\linewidth]{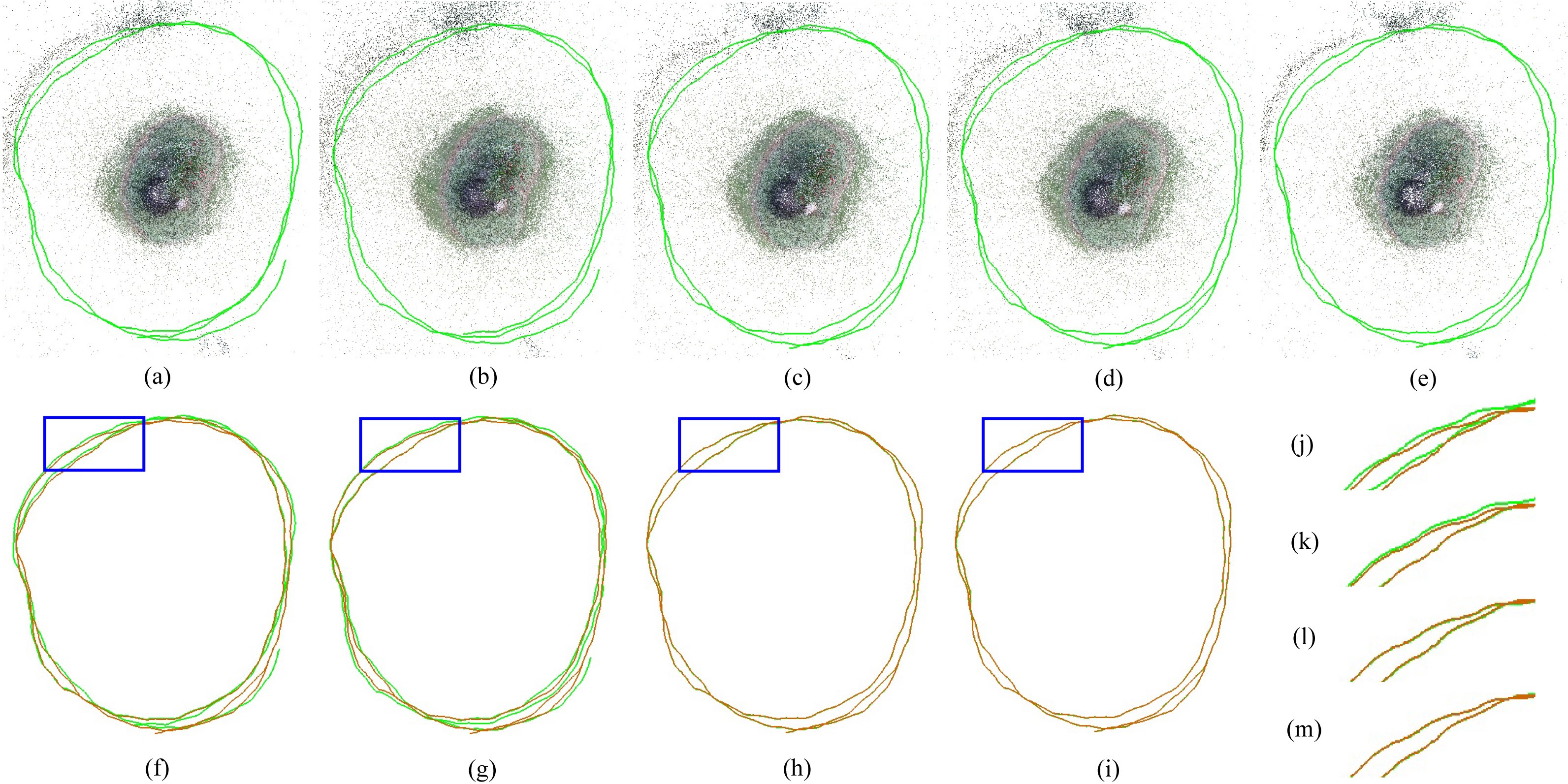}
	\caption{The recovered 3D points~(track length $\ge$ 3) and camera trajectories using feature tracks computed by
		different matching algorithms: (a) C-SIFT; (b) CPT; (c) CPT+NCTM; (d) CPT+KNCTM; (e) BF-SIFT. (f-i) Superimposing the
		camera trajectories~(highlighted in red) in (a-d) to (e). (j-m) Magnified regions of (f)-(i).}
	\label{fig:sfm-comparison}
\end{figure*}

The ``Circle" sequence contains $2,129$ frames. To make computation feasible for a few
prior methods, we select one frame for every 3 consecutive ones, which forms a new
sequence containing 710 frames in total. Table~\ref{tab:statistics} lists the running
time with GPU acceleration. Our consecutive point tracking~(CPT) needs a bit more time
than C-SIFT. But it significantly extends the lifetime of most tracks. With our
non-consecutive track matching, common feature tracks scattered over disjoint
subsequences are connected, further expanding track lifetime. Compared with the
computationally most expensive BF-SIFT, our result~(CPT+NCTM) obtains more long feature
tracks and the computation is much faster. With keyframe-based acceleration, our
non-consecutive track matching time is further significantly reduced~(from 87.6s to 40.9s), without
influencing much matching result. Table~\ref{tab:statistics} lists the average length of
tracks for all tracks with length $\ge 1$. The computed average length is short because
we also take into account unmatched features with track length 1. The quality of SfM results
computed by BF-SIFT, CPT+NCTM and CPT+KNCTM are quite comparable, as shown in Fig.~\ref{fig:sfm-comparison}.

\subsection{Comparison with Other SfM/SLAM Systems}
\begin{table}
	[tb] {
		\caption{Localization error~(RMSE~(m)/Completeness) comparison in KITTI odometry dataset.}
		\scriptsize
		\begin{center}
			\begin{tabular}{|@{\hspace{1mm}}c@{\hspace{1mm}}|@{\hspace{1mm}}c@{\hspace{1mm}}|@{\hspace{1mm}}c@{\hspace{1mm}}|@{\hspace{1mm}}c@{\hspace{1mm}}|@{\hspace{1mm}}c@{\hspace{1mm}}|@{\hspace{1mm}}c@{\hspace{1mm}}|}
				\hline  Seq. & ENFT-SFM & ENFT-SFM & ORB- & VisualSFM & OpenMVG\\
				&          & (Keyframes)&  SLAM  &  (Keyframes) & (Keyframes)     \\
				\hline 00 & 4.58 / 100\% & 4.76 / 100\% & 5.33 & 2.78 / 3.71\% & 5.83 / 0.7\%\\
				\hline 01 & 57.20 / 100\% & 53.96 / 100\% &  X & 52.34 / 12.46\% & 8.79 / 2.08\% \\
				\hline 02 & 28.13 / 100\% & 28.26 / 100\% & 21.28 & 1.77 / 4.53\% & 50.36 / 3.74\% \\
				\hline 03 & 2.82 / 100\% & 2.94 / 100\% & 1.51 & 0.28 / 12.05\% & 3.53 / 8.43\% \\
				\hline 04 & 0.66 / 100\% & 0.66 / 100\% & 1.62 & 0.76 / 23.44\% & 5.14 / 14.06\% \\
				\hline 05 & 2.88 / 100\% & 3.48 / 100\% & 4.85 & 9.77 / 7.42\% & 22.42 / 9.07\% \\
				\hline 06 & 14.24 / 100\% & 14.43 / 100\% & 12.34 & 8.58 / 7.41\% & 3.16 / 3.37\% \\
				\hline 07 & 1.83 / 100\% & 2.03 / 100\% & 2.26 & 3.85 / 7.78\% & 7.75 / 5\% \\
				\hline 08 & 30.74 / 100\% & 28.32 / 100\% & 46.68 & 0.81 / 0.90\% & 17.82 / 2.58\% \\
				\hline 09 & 5.63 / 100\% & 5.88 / 100\% & 6.62 & 0.90 / 4.92\% & 14.26 / 3.36\% \\
				\hline 10 & 19.53 / 100\% & 18.49 / 100\% & 8.80 & 5.70 / 6.05\% & 27.06 / 7.01\% \\
				\hline
			\end{tabular}
		\end{center}
		\label{tab:KITTI-compare}}
\end{table}

\begin{table}
	[tb] {
		\caption{Localization error~(RMSE~(cm)/Completeness) comparison on TUM RGB-D dataset.}
		\scriptsize
		\begin{center}
			\begin{tabular}{|@{\hspace{0.5mm}}c@{\hspace{0.5mm}}|@{\hspace{0.5mm}}c@{\hspace{0.5mm}}|@{\hspace{0.5mm}}c@{\hspace{0.5mm}}|@{\hspace{0.5mm}}c@{\hspace{0.5mm}}|@{\hspace{0.5mm}}c@{\hspace{0.5mm}}|@{\hspace{0.5mm}}c@{\hspace{0.5mm}}|}
				\hline  Sequence & ENFT-SFM & ENFT-SFM & ORB- & VisualSFM & OpenMVG\\
				&          & (Keyframes)&  SLAM  &  (Keyframes) & (Keyframes)     \\
				\hline fr1\_desk & 2.71/99.84\% & 2.96/100\% & 1.69 & 2.74/100\% & X \\
				\hline fr1\_floor & 4.08/96.70\% & 3.93/100\% & 2.99 & 53.11/69.23\% & 0.52/6.92\% \\
				\hline fr1\_xyz & 1.25/100\% & 1.59/100\% & 0.90 & 1.43/100\% & X \\
				\hline fr2\_360\_kidnap & 13.57/91.47\% & 15.31/100\% & 3.81 & 10.08/50.91\% & 5.21/14.55\% \\
				\hline fr2\_desk & 2.43/100\% & 2.27/100\% & 0.88 & 1.79/100\% & 1.38/13.95\% \\
				\hline fr2\_desk\_person & 2.46/100\% & 2.55/100\% & 0.63 & 1.92/100\% & 2.16/97.01\% \\
				\hline fr2\_xyz & 0.81/100\% & 0.73/100\%& 0.30 & 0.71/100\% & 5.74/97.6\% \\
				\hline fr3\_long\_office & 1.21/100\% & 1.44/100\% & 3.45 & 1.15/100\% & 2.94/32.74\% \\
				\hline fr3\_nst\_tex\_far & 3.60/86.58\% & 7.76/100\% & X & 7.29/100\% & 35.64/3.79\% \\
				\hline fr3\_nst\_tex\_near & 1.87/100\% & 1.66/100\% & 1.39 & 1.13/100\% & 3.4/39.13\% \\
				\hline fr3\_sit\_half & 1.50/100\% & 1.55/100\% & 1.34 & 2.30/100\% & 0.68/9.3\% \\
				\hline fr3\_sit\_xyz & 0.84/100\% & 1.39/100\% & 0.79 & 1.28/100\% & 1.03/100\% \\
				\hline fr3\_str\_tex\_far & 0.94/100\% & 0.95/100\% & 0.77 & 2.15/100\% & 1.12/100\% \\
				\hline fr3\_str\_tex\_near & 1.86/100\% & 1.82/100\% & 1.58 & 0.95/100\% & 0.97/19.74\% \\
				\hline fr3\_walk\_half & 2.08/100\% & 2.21/100\% & 1.74 & 1.88/100\% & X \\
				\hline fr3\_walk\_xyz & 1.30/100\% & 1.74/100\% & 1.24 & 1.62/100\% & X \\
				\hline
			\end{tabular}
		\end{center}
		\label{tab:TUM-compare}}
\end{table}

\begin{figure*}[tb]
	\centering
	\includegraphics[width=1.0\linewidth]{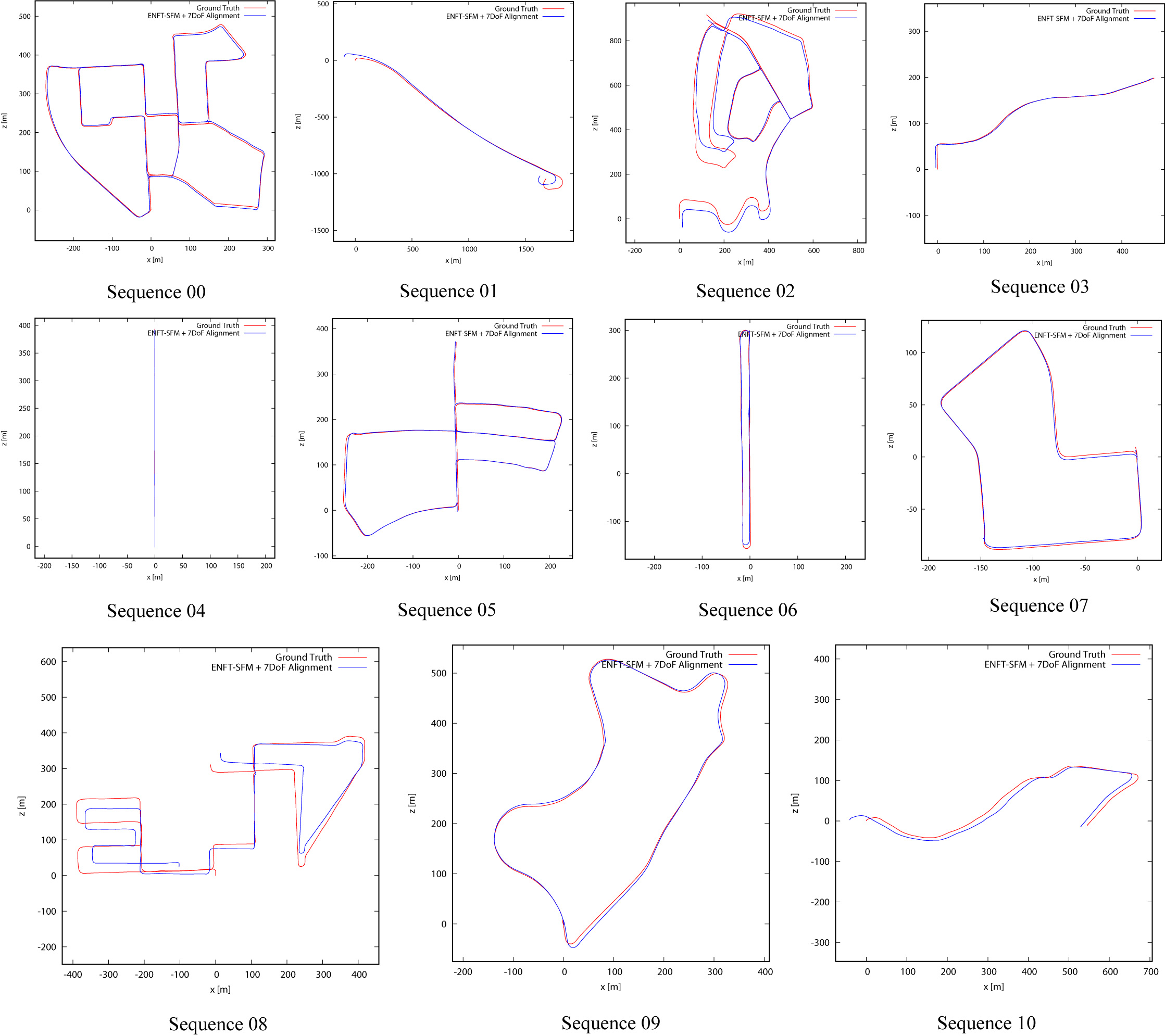}
	\caption{The recovered camera trajectories by ENFT-SFM in KITTI odometry 00-10
		sequences.} \label{fig:KITTI-ENFT}
\end{figure*}

We compare our ENFT-SFM system with state-of-the-art SfM/SLAM systems~(i.e.
ORB-SLAM~\cite{Mur-ArtalMT15}, VisualSFM~\cite{VisualSFM, WuACS11, Wu2013} and OpenMVG~\cite{openMVG})
using our datasets and other public benchmark datasets~(i.e. KITTI odometry
dataset~\cite{Geiger2012CVPR} and TUM RGB-D dataset~\cite{sturm12iros}). Since VisualSFM
and OpenMVG are mainly designed for unordered image datasets, we extract keyframes from
the original sequences as input for VisualSFM and OpenMVG. For fair comparison, our
method processes both original sequences and extracted keyframes for KITTI odometry dataset and TUM RGB-D dataset.

For KITTI and TUM RGB-D datasets, we align recovered camera trajectories and ground
truth by estimating a 7DoF similarity transformation. The RMSE and completeness of camera
trajectories for all methods are listed in Tables~\ref{tab:KITTI-compare} and
\ref{tab:TUM-compare}. ``X'' denotes that the map cannot be accurately initialized or processed. The
recovered camera trajectories of sequences 00-10 from KITTI odometry dataset by
our method are shown in Fig.~\ref{fig:KITTI-ENFT}. Because sequences 01 and 08 do
not contain loops, the drift cannot be corrected, leading to large RMSE.

For ORB-SLAM, we directly quote reported RMSE error of keyframe trajectory in their paper.
Compared with ORB-SLAM, our method achieves comparable results in KITTI odometry dataset. We note
only our method is able to process all sequences~(the camera poses of some frames in TUM RGB-D sequences
are not recovered due to extremely serious motion blur, occlusion or there are not sufficient texture regions). We fix the parameters for both KITTI and TUM RGB-D dataset except for the maximum frame number for each sequence segment. It is set
as $300$ for KITTI odometry dataset and $1,500$ for TUM RGB-D dataset respectively. Since the camera
moves fast in KITTI odometrry dataset, the maximum frame number for each segment should be smaller to reduce
the accumulation error.

\begin{figure}[tb]
    \centering
    \includegraphics[width=1.0\linewidth]{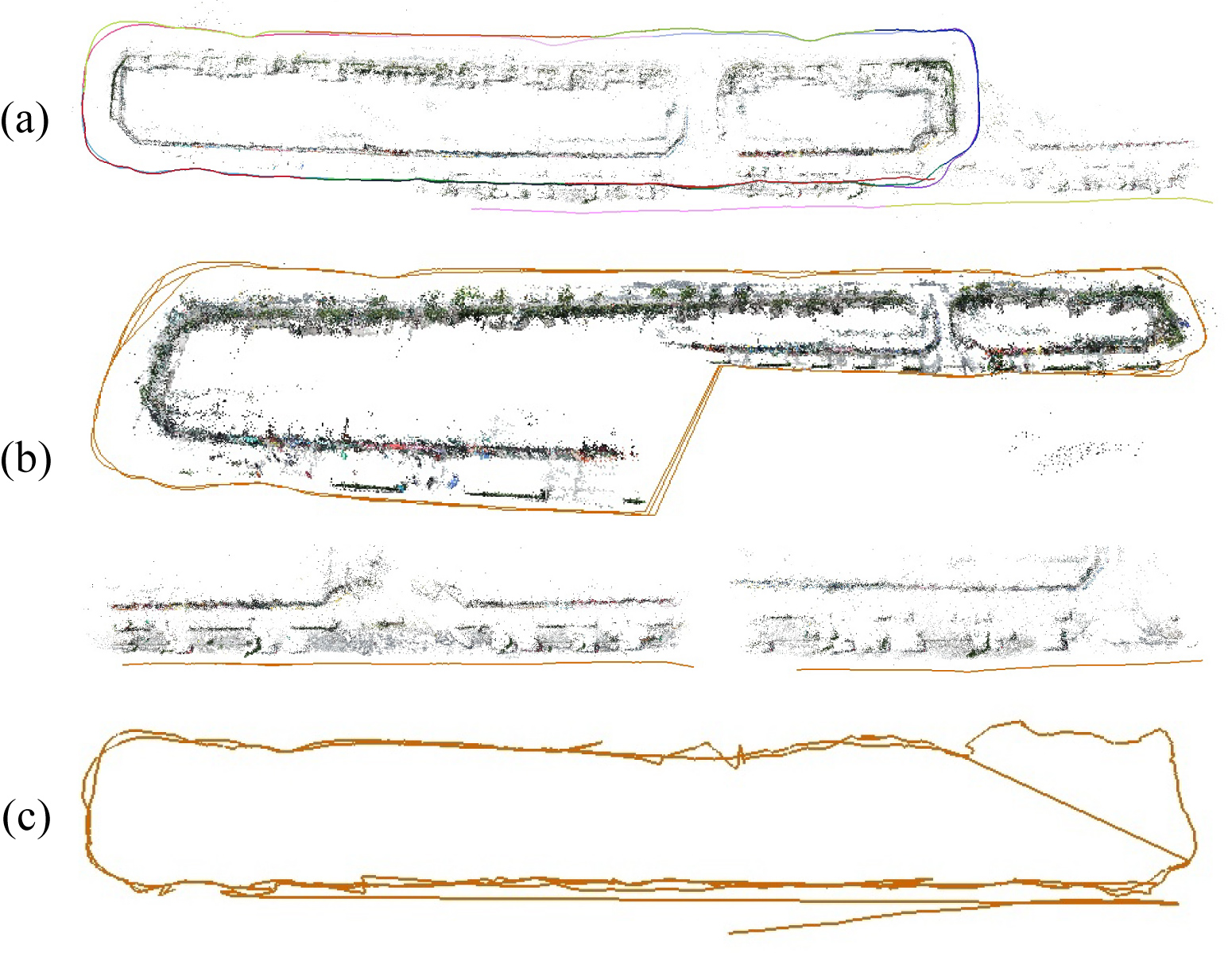}
\caption{Reconstruction comparison on the ``Street'' example. (a) SfM result of ENFT-SFM.
(b) SfM result of VisualSFM, which is separated to 3 individual models. (c) The recovered
camera trajectory by ORB-SLAM.} \label{fig:street-compare}
\end{figure}

\begin{figure}[tb]
    \centering
    \includegraphics[width=1.0\linewidth]{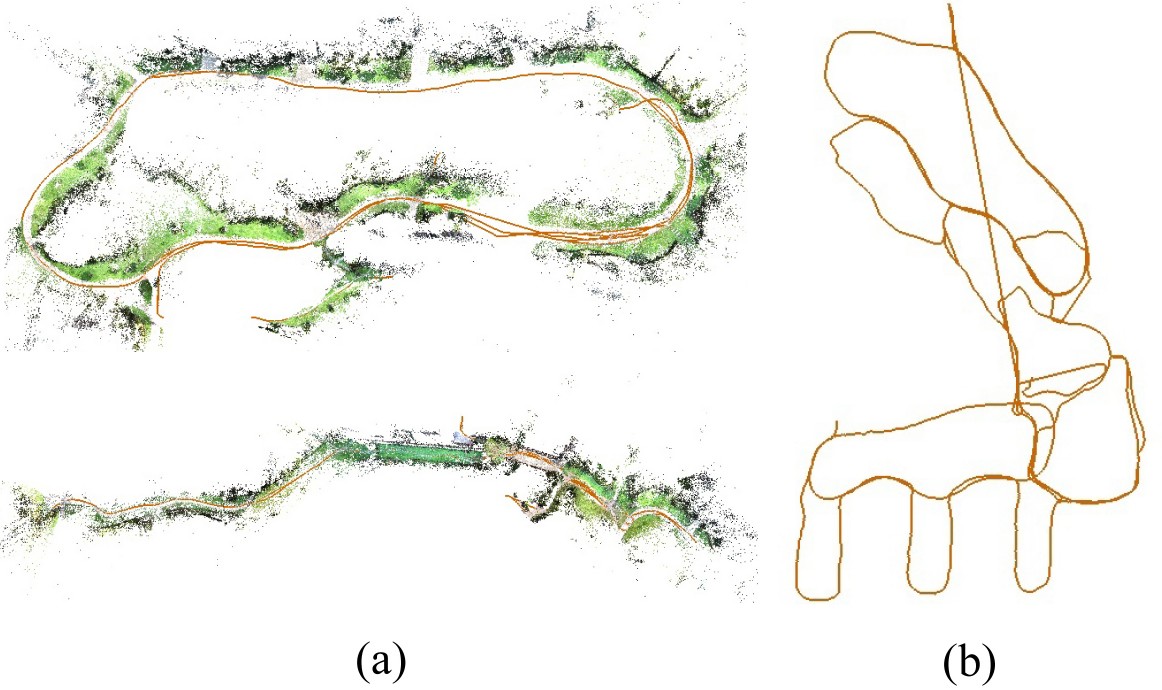}
\caption{Reconstruction comparison on the ``Garden'' example. (a) Two individual models
reconstructed by VisualSFM. The reconstructed SfM result contains 60 individual models.
(b) The recovered camera trajectory by ORB-SLAM.} \label{fig:garden-compare}
\end{figure}

For our multi-sequence data, since ORB-SLAM cannot directly handle multiple sequences, we
constitute multiple sequences into a single sequence by re-ordering the frame index. The
input frame rate is set to 10fps for ORB-SLAM\footnote{We use ORB-SLAM2:
https://github.com/raulmur/ORB\_SLAM2.}. The recovered camera trajectories by ORB-SLAM
are shown in Figs.~\ref{fig:street-compare} and \ref{fig:garden-compare}. The camera poses of
many frames are not recovered due to unsuccessful relocalization. Although some loops
are closed, the optimization is stuck in a local optimum. The reason is twofold. On the
one hand, the matched common features among non-consecutive frames by a traditional
bag-of-words place recognition method~\cite{Galvez-LopezT12} are insufficient for robust
SfM/SLAM. On the other hand, using pose graph
optimization~\cite{StrasdatMD10,KummerleGSKB11} may not sufficiently minimize
accumulation error, and traditional BA is easily stuck in a local optimum if a good starting point is not provided.

VisualSFM does not work that well in KITTI odometry dataset and our long sequences, as shown in
Table~\ref{tab:KITTI-compare} and Figs.~\ref{fig:street-compare} and
\ref{fig:garden-compare}. Note the matching time in our data is overly long for
VisualSFM. We have to use our non-consecutive feature tracking algorithm to get the
feature matching results. The produced SfM results still have the drifting problem and
the whole camera trajectory is easily separated into multiple segments. We thus select
the largest segment for computing RMSE and completeness. One reason for this drifting
problem is the incremental SfM, which may not effectively eliminate accumulated errors.
Another explanation is that sequence continuity/ordering is not completely utilized.
Since the KITTI dataset is captured by an autonomous driving platform and each frame is
only matched to its consecutive frames. Once camera tracking fails in one frame, the
connection between two neighboring subsequences will be broken. In our experiments,
OpenMVG usually performs worse than VisualSFM.

\begin{figure}[tb]
    \centering
    \includegraphics[width=1.0\linewidth]{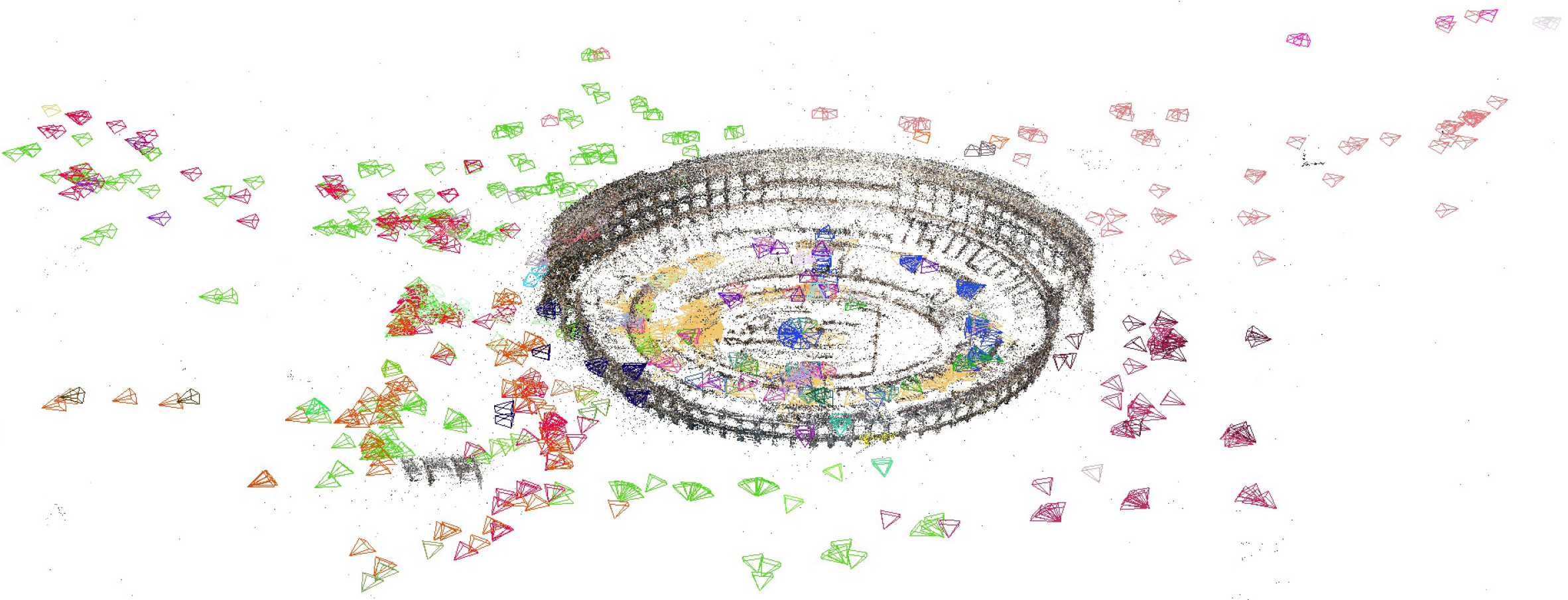}
\caption{The reconstruction result of ``Colosseum'' dataset by our method. Cameras in the same sequence are encoded with the same color.}
    \label{fig:Colosseum}
\end{figure}

\begin{figure*}[tb]
    \centering
    \includegraphics[width=1.0\linewidth]{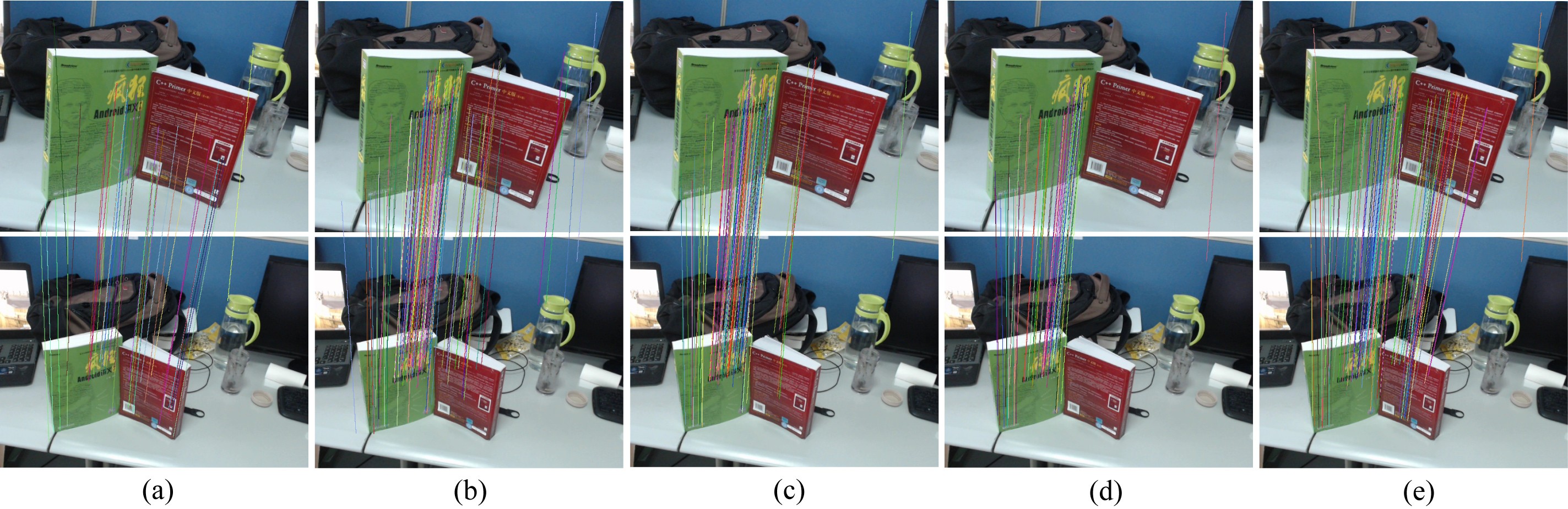}
\caption{Matching result with different $\tau_c$. (a) First-pass matching result. (b-d)
Results of the second-pass matching only using the homography corresponding to the left
green book with $\tau_c = \{0.06, 0.04, 0.02\}$, respectively. The matches that do not
belong to the green book are outliers. (e) Second-pass matching result using all
homographies with $\tau_c = 0.02$. 95 matches are obtained.} \label{fig:param}
\end{figure*}

\subsection{Results on General Image Collections}

Although our segment-based SfM method is originally designed for handling sequences, it can be naturally
extended to work with general image collections. The basic idea is to separate the
unordered image data to a set of sequences according to their common matches.

We first select two images with the maximum number of common features to constitute an
initial sequence. Then we select another image, which has the most common features with
the head or tail frame, and add it into the sequence as the new head or tail. This
process repeats until no image can be added. Then we begin to build another sequence
based on remaining images. For some 3D points that have only one or two corresponding
features in one sequence, we additionally select related images from other sequences to
help estimate the 3D positions.

Fig.~\ref{fig:Colosseum} shows our SfM result on Colosseum dataset~\cite{LiSH10,lou2012matchminer}, which
contains $1,164$ images. We directly use the feature matching result obtained by VisualSFM. Because our current SfM implementation requires that the intrinsic camera parameters and radial distortion are known for each image, we calibrate
the matched feature positions according to the calibrated parameters by VisualSFM. Then
we use our extended segment-based SfM method to estimate camera poses and 3D points. The
processing time of our SfM estimation in a single thread is 125 seconds, which is even shorter
than that of VisualSFM enabling GPU~(269 seconds).

\subsection{Parameter Configuration and Limitation}
The parameters can be easily set in our system because most of them are not sensitive and
use default values. The most important parameter is $\tau_c$, which controls the strength
to mark outliers during feature tracking. A large $\tau_c$ could result in many matches,
and introduce outliers. In our experiments, we conservatively set $\tau_c$ to a small
value $0.02$. By removing a small set of matches, the system becomes reliable for
high-quality SfM. Fig.~\ref{fig:param} shows the matching result with different $\tau_c$.
After the fist-pass matching, 35 matches are obtained. The second-pass matching result
with $\tau_c = 0.06$ is shown in Fig.~\ref{fig:param}(b). A few features that do not
belong to the green book are included. These outliers are removed by using smaller
$\tau_c$ values, as shown in (c) and (d). By setting $\tau_c = 0.02$, almost all outliers
are removed and 95 reliable matches are obtained.

The proposed two-pass matching works best if the scene can be represented by multiple planes.
For a video sequence with dense frames, this condition can be generally achieved because
image transformation between two consecutive frames is small for viable approximation by
one or multiple homographies. We note even if the scene deviates from piecewise
planarity, our second-pass matching still works as rectified images are close to the
target ones. Our method may be not suitable for wide-baseline sparse images where the
number of matches by first-pass matching is too small.

\section{Conclusion and Discussion}
We have presented a robust and efficient non-consecutive feature
tracking (ENFT) method for robust SfM, which consists of two main steps, i.e.,
consecutive point tracking and non-consecutive track matching.
Different from typical sequential matchers, e.g., KLT, we use
invariant features and propose a two-pass matching strategy to
significantly extend the track lifetime and reduce the feature
sensitivity to noise and image distortion. The obtained tracks avail
estimating a match matrix to detect disjointed subsequences with
overlapping views. A new segment-based coarse-to-fine SfM estimation
scheme is also introduced to effectively reduce accumulation error for long sequences.
The presented ENFT-SFM system can handle tracking and registering
large video datasets with limited memory consumption.

Our ENFT method greatly helps SfM, and considers feature tracking on non-deforming
objects by tradition. Part of our future work is to handle dynamic objects. In addition,
although the proposed method is based on SIFT features, there is no limitation to use
other representations, e.g., SURF~\cite{BayETG08} and ORB~\cite{RubleeRKB11}, for further
acceleration.

\section*{Acknowledgements}
We thank Changchang Wu for his kind help in running VisualSFM in our datasets. This work was partially supported by NSF of China~(Nos. 61272048, 61232011), the Fundamental
Research Funds for the Central Universities~(2015XZZX005-05), a Foundation for the Author
of National Excellent Doctoral Dissertation of PR China~(No. 201245), and two grants from
the Research Grants Council of the Hong Kong SAR (Project Nos. 2150760, CUHK417913).

% or





%\bibliographystyle{IEEEtran}

\bibliographystyle{plain}
\bibliography{reference}

\end{document}